\documentclass[journal,letterpaper,10pt]{IEEEtran}
\usepackage{amsmath}
\usepackage{cite}
\usepackage{comment}
\usepackage{graphicx}
\usepackage{amssymb}
\usepackage{subcaption}
\usepackage{xcolor}
\usepackage{booktabs}
\usepackage{multirow}
\usepackage{placeins}
\usepackage{bm}
\usepackage{float}
\IEEEoverridecommandlockouts

\usepackage[hidelinks]{hyperref}
\hypersetup{pdftitle={E2DRL: End-to-End Differentiable Reinforcement Learning for 3D Obstacle Avoidance in Quadruped Robots},
  pdfauthor={Cheng Fan, Bingyang Zhou, Gang Yang, Zixuan Zhuang, Aoqian Zhang, Pengzhi Yang, Xin Liu, Chunzheng Wang, Yan Rui Tan, Fan Shi}}
\makeatletter
\let\NAT@parse\undefined
\makeatother

\title{\LARGE \bf
E2DRL: End-to-End Differentiable Reinforcement Learning for 3D Obstacle Avoidance in Quadruped Robots
}

\author{Cheng~Fan\textsuperscript{1,*}, Bingyang~Zhou\textsuperscript{1,*}, Gang~Yang\textsuperscript{1}, Zixuan~Zhuang\textsuperscript{1}, Aoqian~Zhang\textsuperscript{1}, Pengzhi~Yang\textsuperscript{1},\\
Xin~Liu\textsuperscript{1}, Chunzheng~Wang\textsuperscript{1}, Yan~Rui~Tan\textsuperscript{1}, and Fan~Shi\textsuperscript{1,\textdagger}%
\thanks{\textsuperscript{*}Equal contribution. \textsuperscript{\textdagger}Corresponding author.}%
\thanks{\textsuperscript{1}Department of Electrical and Computer Engineering, National University of Singapore, Singapore.}%
}

\IEEEaftertitletext{\vspace{-2\baselineskip}}

\begin{document}

\pagestyle{empty}
\begingroup
\clubpenalty=10000
\widowpenalty=10000
\setlength{\textfloatsep}{10pt plus 2pt minus 2pt}
\setlength{\dbltextfloatsep}{10pt plus 2pt minus 2pt}
\setlength{\floatsep}{8pt plus 2pt minus 2pt}
\setlength{\dblfloatsep}{8pt plus 2pt minus 2pt}

\maketitle
\thispagestyle{empty}

\begin{abstract}

Differentiable reinforcement learning shows strong promise for robot policy learning, but its application to legged locomotion has remained limited to blind walking or simulated 2D tasks. Extending it to end-to-end visual obstacle avoidance is challenging because hard collision constraints and early termination upon contact corrupt first-order gradients, trapping optimizers in local minima. We present E2DRL, which addresses this with a soft differentiable obstacle-interaction formulation: penetration penalties across multiple body keypoints with velocity-conditioned stop-gradient shaping to provide the first-order training signals for 3D obstacle avoidance. To realize a deployable perception-control system, a privileged teacher trained via Soft Analytic Policy Optimization (SAPO) in a differentiable simulator is distilled into a recurrent depth-based student via DAgger, then blended with a base locomotion policy using a terrain-aware gating module for deployment. In simulation, E2DRL achieves higher sample efficiency than PPO and demonstrates superior velocity-tracking performance and 3D obstacle-avoidance capability compared with both PPO and Duality-based DCBF NMPC. We present the first sim-to-real validation of end-to-end differentiable RL for 3D obstacle avoidance on a Unitree Go2 in static and dynamic environments.

\end{abstract}

\section{INTRODUCTION}

Quadruped robots have shown strong potential for operation in human-centered environments \cite{tranzatto2022cerberus,mattamala2022efficient}, where obstacle avoidance often extends beyond planar navigation. In cluttered scenes, a velocity-commanded robot may encounter obstacles with diverse heights and geometries, such as tables, shelves, chairs, and suspended boards. To navigate such environments effectively, the robot must not only track commanded motion, but also adapt its body posture and locomotion strategy in three dimensions, e.g., bypassing obstacles laterally or lowering its body to crawl underneath them. Enabling such command-conditioned 3D obstacle avoidance is therefore important for the practical deployment of quadruped robots \cite{miki2024learning,cheng2024extreme,margolis2023walk,sun2020path,liao2023walking}.

Recent studies have explored quadrupedal locomotion in complex environments through both optimization-based and learning-based approaches \cite{liao2023walking,rudin2022learning,lee2020learning,miki2022learning,kumar2021rma,miki2024learning,cheng2024extreme}. Traditional optimization-based methods, such as Duality-based DCBF NMPC \cite{liao2023walking}, incorporate safety constraints into online trajectory optimization for safe quadrupedal locomotion in narrow spaces. However, their reliance on online constrained optimization and simplified model representations makes it difficult to simultaneously avoid challenging obstacles while maintaining responsive velocity tracking. Model-free reinforcement learning offers an alternative, but typically treats the simulator as a black box and estimates gradients only from sampled returns \cite{suh2022differentiable}, requiring large amounts of interaction data in contact-rich obstacle-avoidance tasks.

Differentiable simulation has recently emerged as a promising alternative for more efficient policy learning \cite{suh2022differentiable,freeman2021brax,howell2022dojo}. Prior work has also shown that differentiable physics can improve motion performance. Notably, Zhang et al. \cite{zhang2024back} demonstrate high-speed drone navigation through complex environments with both static and dynamic obstacles, achieving substantial gains in speed and success rate. In addition, Schwarke et al. \cite{schwarke2024learning} achieve robust proprioceptive quadrupedal locomotion by proposing a novel differentiable contact model, enabling seamless sim-to-real transfer with remarkable sample efficiency. These results suggest that differentiable physics can improve both optimization efficiency and control performance. We therefore seek to bring these advantages to quadrupedal 3D obstacle avoidance.

Nevertheless, applying differentiable RL to quadrupedal obstacle avoidance remains challenging. Intermittent contacts, hard collision constraints, and early termination induce severe non-smoothness and gradient discontinuities. Consequently, first-order optimization methods become highly susceptible to suboptimal local minima. Existing differentiable RL methods for quadruped robots mainly address blind locomotion or simulated 2D tasks \cite{song2024learning,luo2025residual}, leaving the integration of differentiable learning into deployable visuomotor policies for 3D obstacle avoidance unexplored. In this setting, the central challenge is not simply to add an obstacle-avoidance cost, but to design a training surrogate that keeps obstacle interactions informative for first-order policy optimization while discouraging nonphysical penetration.

In this letter, we present \textit{E2DRL}, an end-to-end differentiable reinforcement learning framework for quadrupedal 3D obstacle avoidance under velocity commands. The framework trains a privileged teacher with SAPO in a differentiable simulator, distills it into a recurrent depth-based student via DAgger, and blends the student with a proprioception-based locomotion policy through a terrain-aware gating module. A key innovation of this work is a soft differentiable obstacle-interaction formulation, used as a training-time surrogate rather than a physical contact model, which allows simulated penetration and applies differentiable multi-keypoint penalties with velocity-conditioned stop-gradient shaping. The gating module preserves nominal locomotion in free space while triggering active avoidance near obstacles. Experiments show that E2DRL achieves higher sample efficiency than PPO during training, while outperforming both Duality-based DCBF NMPC and PPO in velocity tracking and traversal of challenging 3D obstacles.

The main contributions of this paper are summarized as follows:
\begin{enumerate}
    \item We propose a soft differentiable obstacle-interaction formulation for contact-rich quadrupedal obstacle avoidance, which addresses the local-optima and gradient-truncation issues of hard obstacle formulations by penalizing continuous penetration of multiple body keypoints with velocity-conditioned stop-gradient shaping, enabling stable first-order optimization across overhanging and ground obstacle configurations.
    \item We propose a deployment-oriented differentiable end-to-end obstacle avoidance framework that transfers differentiable RL skills for crawling and bypassing to a recurrent depth-based student policy. 
    Furthermore, to improve velocity tracking in obstacle-free regions, we introduce a gating module that seamlessly integrates the student policy with a nominal locomotion policy.
    \item  We present the first sim-to-real validation of an end-to-end differentiable RL policy for overhanging and ground obstacle avoidance on a physical quadruped robot in both static and dynamic real-world environments. 
\end{enumerate}

\section{Related Work}

\subsection{Learning-based Quadrupedal Locomotion}

Deep reinforcement learning has significantly improved quadrupedal locomotion over diverse terrains. Rudin et al. \cite{rudin2022learning} demonstrated that basic walking skills can be acquired within minutes through massively parallel training, while Hwangbo et al. \cite{hwangbo2019learning} showed that reinforcement learning can produce agile motor skills such as rapid direction changes and recovery from external disturbances. 

Building on these advances, recent studies have extended quadrupedal locomotion to more challenging navigation scenarios involving obstacles and confined spaces. Miki et al. \cite{miki2022learning} enabled locomotion in the wild. He et al. \cite{he2024agile} achieved high-speed locomotion with obstacle avoidance in cluttered environments. Zhuang et al. \cite{zhuang2023robot} demonstrated parkour-style behaviors that require coordinated whole-body control. In contrast to these prior works, our work focuses on end-to-end 3D obstacle avoidance with velocity-command tracking.

\subsection{Differentiable Reinforcement Learning}

Gradient-based policy optimization through differentiable simulation enables analytical policy gradients with lower variance than sampling-based methods~\cite{mozer2013focused,suh2022differentiable,freeman2021brax,howell2022dojo}. However, discontinuous contact dynamics create non-smooth optimization landscapes that cause BPTT-based methods to converge to poor local optima. Xu et al.~\cite{xu2022accelerated} proposed SHAC to mitigate this via short-horizon differentiable rollouts combined with an actor-critic baseline, and Xing et al.~\cite{xing2024stabilizing} further introduced SAPO with maximum entropy regularization~\cite{haarnoja2018soft} to smooth the landscape. Although existing differentiable methods have demonstrated success in robust blind locomotion and simulated 2D tasks \cite{song2024learning,schwarke2024learning,luo2025residual}, the application to 3D obstacle avoidance remains unexplored. This is primarily due to the highly non-smooth optimization landscapes induced by hard constraints, as well as gradient truncation resulting from early termination. We bridge this gap by extending SAPO with a soft differentiable obstacle-interaction formulation.

\begin{figure*}[t]
\centering
\includegraphics[width=1.0\linewidth]{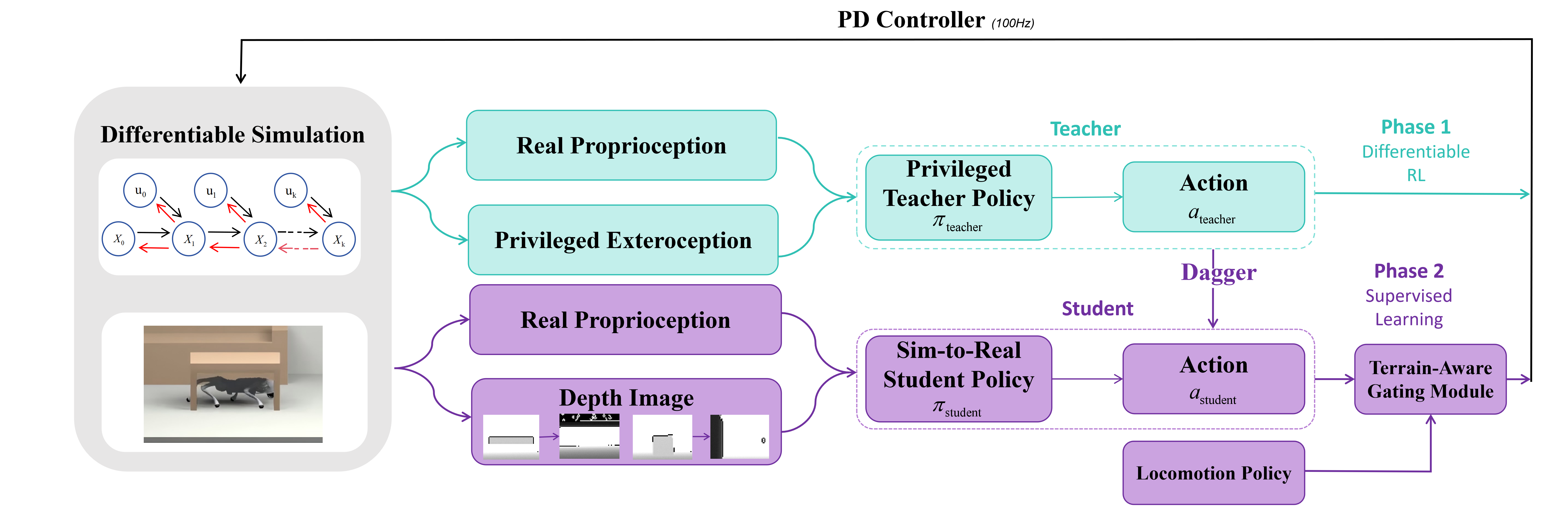}
\caption{\textbf{Overview of E2DRL.} We first train a privileged teacher with SAPO using the proposed soft differentiable obstacle-interaction formulation. The teacher is then distilled into a recurrent depth-based student via DAgger. For deployment, a terrain-aware gating module blends the student policy with a nominal locomotion policy.}
\label{fig:overview}
\end{figure*}

\section{PROBLEM FORMULATION}

We address vision-based 3D obstacle avoidance for a quadruped robot given desired velocity commands. The robot must traverse environments with overhanging obstacles that require body-height adaptation and lateral ground obstacles that require bypassing.

\textbf{Observation Space:}
The robot observation comprises two modalities. The proprioceptive state $s_t^{\text{proprio}} \in \mathbb{R}^{45}$ includes base angular velocity, projected gravity vector, base velocity commands, joint positions, joint velocities, and last actions. The privileged exteroceptive state $s_t^{\text{priv-extero}} \in \mathbb{R}^{14}$ includes base position, base linear velocity, Euler angles, obstacle configurations and geometric obstacle parameters (clearance height, lateral width and longitudinal depth). During training, the full privileged observation $s_t^{\text{priv}} = (s_t^{\text{proprio}},\, s_t^{\text{priv-extero}})$ is available to the teacher policy. For deployment, the privileged exteroceptive component is replaced by a depth image $I_t \in \mathbb{R}^{64\times64}$ from the onboard camera, yielding the student observation $s_t^{\text{stu}} = (s_t^{\text{proprio}},\, I_t)$.

\textbf{Action Space:}
The action $a_t \in \mathbb{R}^{12}$ specifies target joint positions for the 12 actuated joints, tracked by PD controllers with gains $K_p = 40$ and $K_d = 1$.

\section{Method}

\subsection{System Overview}
Our approach is a two-stage learning framework for 3D obstacle avoidance using only onboard perception, as illustrated in Fig.~\ref{fig:overview}. It bridges the sample-efficient First-Order Policy Gradient (FoPG) method with vision-based real-world deployment.

In the first stage (Sec.~\ref{sec:teacher}), a privileged teacher policy is trained in a differentiable simulator via Soft Analytic Policy Optimization (SAPO). To improve optimization robustness in 3D obstacle avoidance scenarios, we incorporate a stable frictional contact model and introduce a soft differentiable obstacle-interaction formulation.

In the second stage (Sec.~\ref{sec:student}), DAgger~\cite{ross2011reduction} distills the teacher into a vision-based student policy that relies solely on onboard depth images and proprioception.

For deployment (Sec.~\ref{sec:deployment}), a terrain-aware gating module dynamically interpolates between the original proprioceptive locomotion policy and the obstacle-avoidance student policy. The gating module is trained separately and is also used during distillation data collection. This design preserves the student's end-to-end observation-to-action mapping while integrating it with the locomotion policy to preserve nominal locomotion capability.

\subsection{Privileged Teacher Policy via Differentiable RL}
\label{sec:teacher}

We model the robot-environment interaction as a Markov Decision Process with differentiable dynamics:
\begin{equation}
    s_{t+1} = f(s_t, a_t)
\end{equation}
where $f: \mathcal{S} \times \mathcal{A} \rightarrow \mathcal{S}$ is a differentiable transition function enabling analytical gradients $\partial s_{t+1}/\partial s_t$ and $\partial s_{t+1}/\partial a_t$ to be backpropagated through the simulation rollout. The teacher policy is optimized via SAPO to maximize the $k$-step objective:
\begin{equation}
    \tilde{r}_t
    \triangleq
    r_t + \alpha \mathcal{H}_{\pi}(a_t \mid s_t)
    \label{eq:entropy_augmented_reward}
\end{equation}
\begin{equation}
    L_{t:t+k}
    =
    \sum_{l=0}^{k-1} \gamma^l \tilde{r}_{t+l}
    +
    \gamma^k V_{\mathrm{soft}}(s_{t+k})
    \label{eq:compact_soft_k_step_return}
\end{equation}
where $r_t$ is the environment reward at timestep $t$, $\alpha$ is the entropy temperature balancing reward maximization and exploration, $\mathcal{H}_{\pi}(a_t \mid s_t)$ denotes the policy entropy, $\gamma \in [0,1)$ is the discount factor, and $V_{\text{soft}}(\bm{s}_{t+k})$ is the soft value function at the terminal state.

For the frictional contact model, we employ the differentiable contact formulation from Geilinger et al. (2020) \cite{geilinger2020add} instead of the standard friction model used in NVIDIA Warp \cite{macklin2022warp}. The key advantage of this approach lies in its gradient stability when contact velocities approach zero. The contact dynamics are described by:
\begin{equation}
    \mathbf{f}_c = (-k_n + k_d \dot{d}) \min(d, 0) \mathbf{n}
    \label{eq:fc}
\end{equation}
\begin{equation}
    \mathbf{f}_t = -\frac{\mathbf{v}_t}{\|\mathbf{v}_t\|} \min(k_t \|\mathbf{v}_t\|, \mu \|\mathbf{f}_c\|)
    \label{eq:ft}
\end{equation}
where $\mathbf{f}_c$ and $\mathbf{f}_t$ are the normal and tangential contact forces, $d$ is the contact penetration depth, $\dot{d}$ its rate, $\mathbf{n}$ the contact normal vector, $\mathbf{v}_t$ the relative tangential velocity, and $k_n$, $k_d$, $k_t$, and $\mu$ denote the normal stiffness, normal damping coefficient, tangential damping coefficient, and friction coefficient, respectively. At $\mathbf{v}_t=\mathbf{0}$, we define $\mathbf{f}_t=\mathbf{0}$ by continuous extension.

Directly training a visual policy via differentiable RL is computationally prohibitive and prone to poor local optima, often resulting in undesirable behaviors such as the robot getting stuck near obstacles or failing to avoid collisions. Therefore, we first train a teacher policy $\pi_{\text{teacher}}(a_t \mid s^{\text{priv}}_t)$ using the privileged observation $s^{\text{priv}}_t$. The policy is parameterized as an MLP.

The proposed obstacle interaction is a differentiable training surrogate rather than a model of physical obstacle contact: it allows the robot's body to penetrate obstacles in simulation while applying a continuous differentiable penalty to sampled body keypoints. Instead of terminating episodes at obstacle contact, we keep the rollout active and provide dense first-order training signals for both crawling under overhanging obstacles and bypassing ground obstacles.

To achieve 3D obstacle avoidance, a differentiable penetration penalty is computed across a set of strategically selected robot keypoints, including the base center of mass, joints, foot end-effectors, and two points at the front and rear of the base, all of which are fully differentiable with respect to the robot's state.
Let $\delta_i^{m}
=
\left|\left|p_i^{m}\right|-d_{\text{obstacle}}^{m}\right|
\cdot \mathbf{1}[x_i\in\mathcal{X}_{\text{obs}}]
\cdot \mathbf{1}[\mathcal{C}^{m}_i]$ denote the penetration depth of the $i$-th keypoint. Here, $p_i^{m}$ is the keypoint coordinate in the obstacle frame, $d_{\text{obstacle}}^{m}$ is the corresponding safe boundary, and $\mathcal{C}^{m}_i$ is the boundary-violation condition, where $m\in\{\text{oh},\text{gr}\}$ indexes the overhanging or ground obstacle. The penalty is activated only when a keypoint penetrates the obstacle. Consequently, collision-free traversal yields $\delta_i^{m}=0$. The penetration reward is then defined as:
\begin{equation}
r^{\text{pen}} = 
\begin{cases} 
-\sum_{i} \delta_i^{m} \cdot e^{\frac{\text{sg}(v_x)}{\sigma}} + k, & \text{if } \sum_{i} \delta_i^{m} > 0 \\
0, & \text{otherwise}.
\end{cases}
\label{eq:penetration_reward}
\end{equation}
where $\sigma$ is a scaling constant, $v_x$ is the forward velocity, and $k$ is a negative constant. The operator $\text{sg}(\cdot)$ denotes the stop-gradient operation. By detaching $v_x$ from the computational graph, the velocity term acts solely as a scaling factor for the penalty magnitude. $\text{sg}(\cdot)$ therefore introduces no gradient along $v_x$ and does not push the policy toward a lower forward velocity, which preserves velocity tracking. The velocity-conditioned shaping as a whole instead discourages the robot from traversing an obstacle at high speed while penetration occurs. Finally, the negative constant $k$ provides a discriminatory signal to the critic network, allowing it to clearly discern whether the robot has successfully avoided obstacles.

\subsection{Vision-based Student Policy Distillation}
\label{sec:student}
In the second stage, we distill the privileged teacher into a vision-based student policy $\pi_{\text{student}}(a_t | s^{\text{stu}}_t)$ that can be deployed on physical hardware.

The student network utilizes a Convolutional Neural Network (CNN) feature extractor. The flattened visual features are concatenated with the proprioceptive state and fed into a Gated Recurrent Unit (GRU) to capture the spatio-temporal dynamics of the obstacles. Finally, an MLP maps the GRU output to the 12-dimensional joint position targets. Layer sizes, hidden-state handling, and the remaining implementation details are reported in the supplementary material.

We employ DAgger \cite{ross2011reduction} to train the student. The training distribution is a mixture of teacher and student rollouts:
\begin{equation}
    \rho \triangleq \beta \rho_{\text{teacher}} + (1 - \beta) \rho_{\text{student}}
\end{equation}
and the distillation loss minimizes the L2 loss between teacher and student actions:
\begin{equation}
    L_{\text{distill}} = \mathbb{E}_{\sigma_t \sim \rho} \left[\left\|\pi_{\text{teacher}}(\sigma_t^{\text{teacher}}) - \pi_{\text{student}}(\sigma_t^{\text{student}})\right\|_2\right] \label{eq:dagger_loss}
\end{equation}
where $\rho_{\text{teacher}}$ and $\rho_{\text{student}}$ are the distributions induced by teacher and student rollouts, and $\beta$ is the mixing probability. Unlike previous approaches that require multiple GPUs for distillation \cite{zhuang2023robot,chen2025multi}, our method decouples data collection and training into alternating phases, enabling single-GPU training. The mixing probability $\beta$ is predefined and decays linearly from one to zero over the course of training, progressively shifting control from the teacher to the student policy. At each iteration, the collected data is aggregated with all previous data, enabling training with the Follow-The-Leader (FTL) algorithm. A denser distribution of the predefined mixing probability $\beta$ leads to better distillation performance.

\subsection{Training Details}

\textbf{Curriculum Learning:}
Training follows a two-stage curriculum. The quadruped robot is first trained for basic velocity-tracking locomotion by requiring it to reach a target distance threshold. After achieving consistent success, training advances to the 3D obstacle avoidance stage, where the robot is exposed to generated obstacle configurations with progressively increasing difficulty. Difficulty is increased once the robot completes five consecutive collision-free traversals.

\textbf{Reward Design:}
The per-step reward $r_t$ is a weighted sum of the terms detailed in Table~\ref{tab:reward_terms}. Specifically, $r^{\text{pen}}_{\text{oh}}$ and $r^{\text{pen}}_{\text{gr}}$ penalize the same penetration depth $\delta_i^{m}$ accumulated over the body keypoints indexed by $i$ that lie inside the obstacle region $\mathcal{X}_{\text{obs}}$. Furthermore, $\text{sg}(\cdot)$ represents the stop-gradient operator, $\sigma$ is set to 0.5 and the term $k$ serves as a conditional penalty that equals $-0.3$. Only $r^{\mathrm{pen}}$ is included in the weighted per-step reward. Finally, $l$ and $j$ denote the indices of the feet and joints, respectively.

\begin{table}[!t]
\centering
\caption{Reward terms used during training.}
\label{tab:reward_terms}
\renewcommand{\arraystretch}{1.2}
\resizebox{\columnwidth}{!}{%
\begin{tabular}{l l l c}
\toprule
\textbf{Symbol} & \textbf{Description} & \textbf{Formulation} & \textbf{Weight} \\
\midrule
\multicolumn{4}{l}{\textit{Task tracking}} \\
$r^{\text{lin}}$      & Linear velocity tracking      & $-\sqrt{(v_x^{\text{cmd}}-v_x)^2 + (v_y^{\text{cmd}}-v_y)^2}$ & $2.0$ \\
$r^{\text{ang}}$      & Yaw rate tracking             & $-\sqrt{(\omega_z^{\text{cmd}}-\omega_z)^2}$ & $1.0$ \\
\midrule
\multicolumn{4}{l}{\textit{Stability}} \\
$r^{\text{ang}_{xy}}$ & Roll/pitch angular velocity   & $-\sum_{i\in\{x,y\}} \omega_i^2$ & $0.05$ \\
$r^{\text{height}}$   & Base height tracking          & $-|h_{\text{base}} - 0.34|$ & $4.0$ \\
\midrule
\multicolumn{4}{l}{\textit{Smoothness \& effort}} \\
$r^{\tau}$            & Joint torque penalty          & $-\sum_{j=1}^{12}(K_p(a_j{-}q_j){-}K_d\dot q_j)^2$ & $3{\times}10^{-3}$ \\
$r^{\ddot{q}}$        & Joint acceleration penalty    & $-\sum_{j=1}^{12}((\dot q_j^{t}{-}\dot q_j^{t-1})/\Delta t)^2$ & $1{\times}10^{-7}$ \\
$r^{\Delta a}$        & Action-rate penalty           & $-\sum_{j=1}^{12}(a_j^{t}{-}a_j^{t-1})^2$ & $0.01$ \\
\midrule
\multicolumn{4}{l}{\textit{Gait shaping}} \\
$r^{\text{foot}}$     & Foot-height sinusoidal tracking & $\sum_{l=1}^{4}\exp({-}|z_l{-}z_l^{*}|/\sigma)$& $0.1$ \\
$r^{\text{air}}$      & Max air-time penalty          & $-\sum_{l}(t_l^{\text{air}}{-}t_{\max})\cdot\mathbf{1}[t_l^{\text{air}}{>}t_{\max}]$& $2.0$ \\
\midrule
\multicolumn{4}{l}{\textit{Obstacle-aware penetration}} \\
$r^{\text{pen}}_{\text{oh}}$  & Soft penetration, overhanging obstacle & $-\sum_{i}\left|\left|p_i^{\text{oh}}\right|{-}d_{\text{obstacle}}^{\text{oh}}\right|\cdot\mathbf{1}[x_i\!\in\!\mathcal{X}_{\text{obs}}]\cdot\mathbf{1}[\mathcal{C}^{\text{oh}}_i]$& $40$ \\
$r^{\text{pen}}_{\text{gr}}$ & Soft penetration, ground obstacle  & $-\sum_{i}\left|\left|p_i^{\text{gr}}\right|{-}d_{\text{obstacle}}^{\text{gr}}\right|\cdot\mathbf{1}[x_i\!\in\!\mathcal{X}_{\text{obs}}]\cdot\mathbf{1}[\mathcal{C}^{\text{gr}}_i]$& $10$ \\
$r^{\text{pen}}$                 & Velocity-scaled penetration & $\left(-\sum_i \delta_i^{m} \cdot e^{\mathrm{sg}(v_x)/\sigma} + k\right) \cdot \mathbf{1}[\sum_i \delta_i^{m} {>} 0]$& $1.0$ \\
\midrule
\multicolumn{4}{l}{\textit{Terminal \& alive}} \\
$r^{\text{term}}$     & Early-termination penalty     & $-10$ on fall / out-of-corridor & $1.0$ \\
$r^{\text{alive}}$    & Survival bonus                & $+1$ per step & $2.5$ \\
\bottomrule
\end{tabular}%
}
\end{table}

\textbf{Deployment:}
\label{sec:deployment}
To mitigate the degradation in forward-velocity tracking caused by distillation, we introduce a gating module that improves student velocity tracking accuracy in obstacle-free segments. Since raw depth alone cannot reliably identify obstacles that require avoidance, the terrain-aware gating module learns to estimate their proximity directly from exteroceptive visual observations. Specifically, it triggers a binary switch from the base locomotion policy to the obstacle-avoidance student policy when the visually estimated distance to the obstacle surface falls below $0.6\text{ m}$. The base locomotion policy is trained in the NVIDIA Warp differentiable simulator. The observation of the base locomotion policy comprises only the proprioceptive state $s^{\text{proprio}}_t \in \mathbb{R}^{45}$. To avoid abrupt action discontinuities during switching, the outputs of the locomotion and obstacle-avoidance policies are interpolated over a short transition window of 30 steps as
$a_t = (1-\lambda_t)a_t^{\mathrm{base}} + \lambda_t a_t^{\mathrm{student}}$,
where $\lambda_t$ smoothly ramps between 0 and 1, with the direction of the transition dictated by the gating module. The gating module adopts the same network architecture as the student policy and is trained using depth observations collected from both Isaac Gym and NVIDIA Warp simulation and real-world data to bridge the sim-to-real gap.
Although trained separately, the gating module performs policy switching and action blending during distillation data collection.

\FloatBarrier

\section{Experiments}

We evaluate the proposed framework in both simulation and real-world experiments. In simulation (Sec.~\ref{sim}), we evaluate the differentiable reinforcement learning approach by comparing its training sample efficiency against PPO, while benchmarking its velocity-tracking performance and 3D obstacle-avoidance capability against both PPO and the Duality-based DCBF NMPC baseline carefully modified with 3D CBF constraints. We further conduct ablation experiments to analyze the contribution of key design components. In real-world experiments (Sec.~\ref{real}), we further examine the sim-to-real transferability of the vision-based policy on a Unitree Go2 platform in indoor and outdoor environments with both static and dynamic obstacles. 

\begin{figure}[t]
    \centering
    \subfloat[Top-view trajectory]{
        \includegraphics[width=0.45\linewidth]{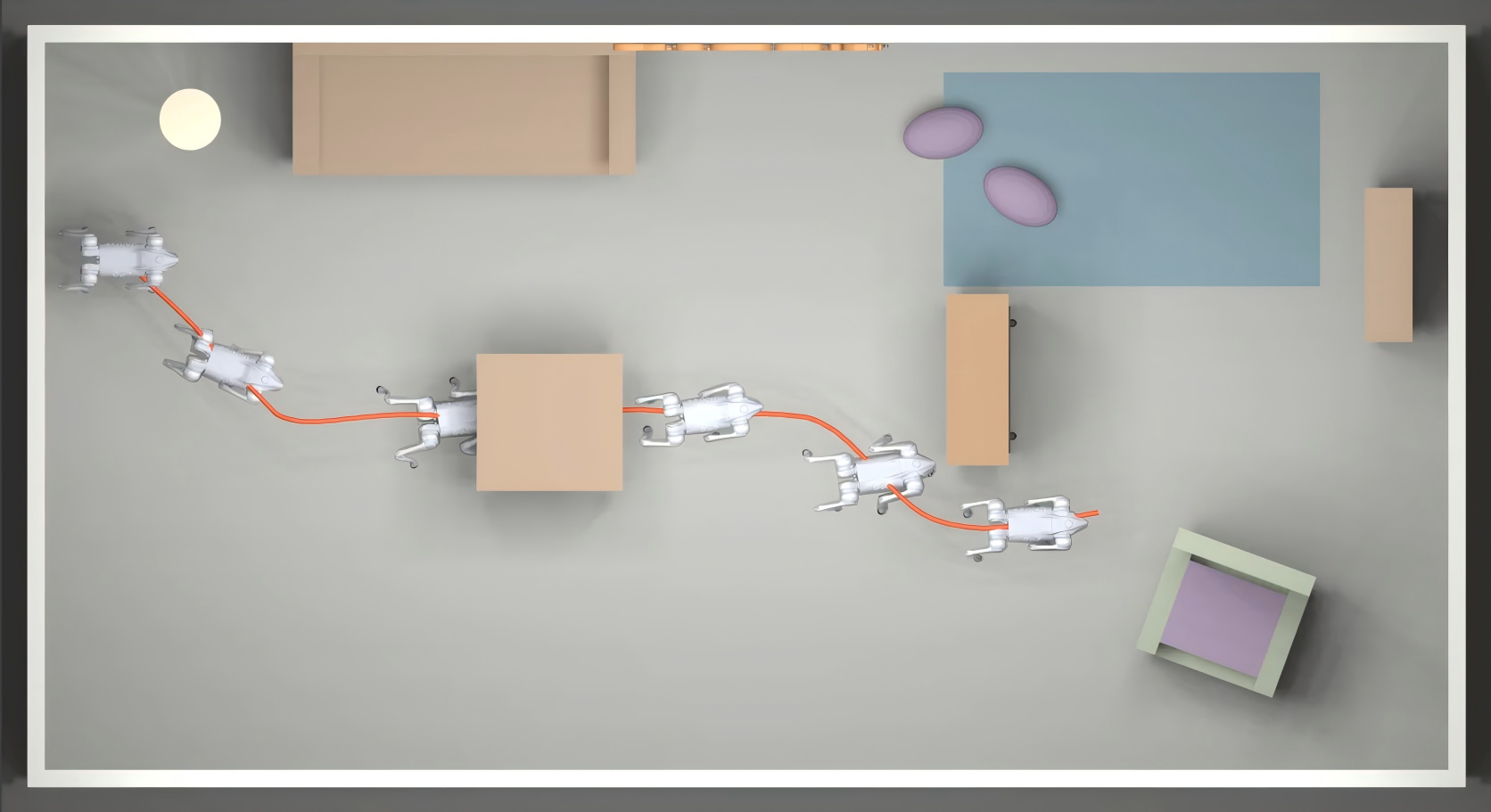}
        \label{fig:top}
    }
    \hfill
    \subfloat[Side-view trajectory]{
        \includegraphics[width=0.45\linewidth]{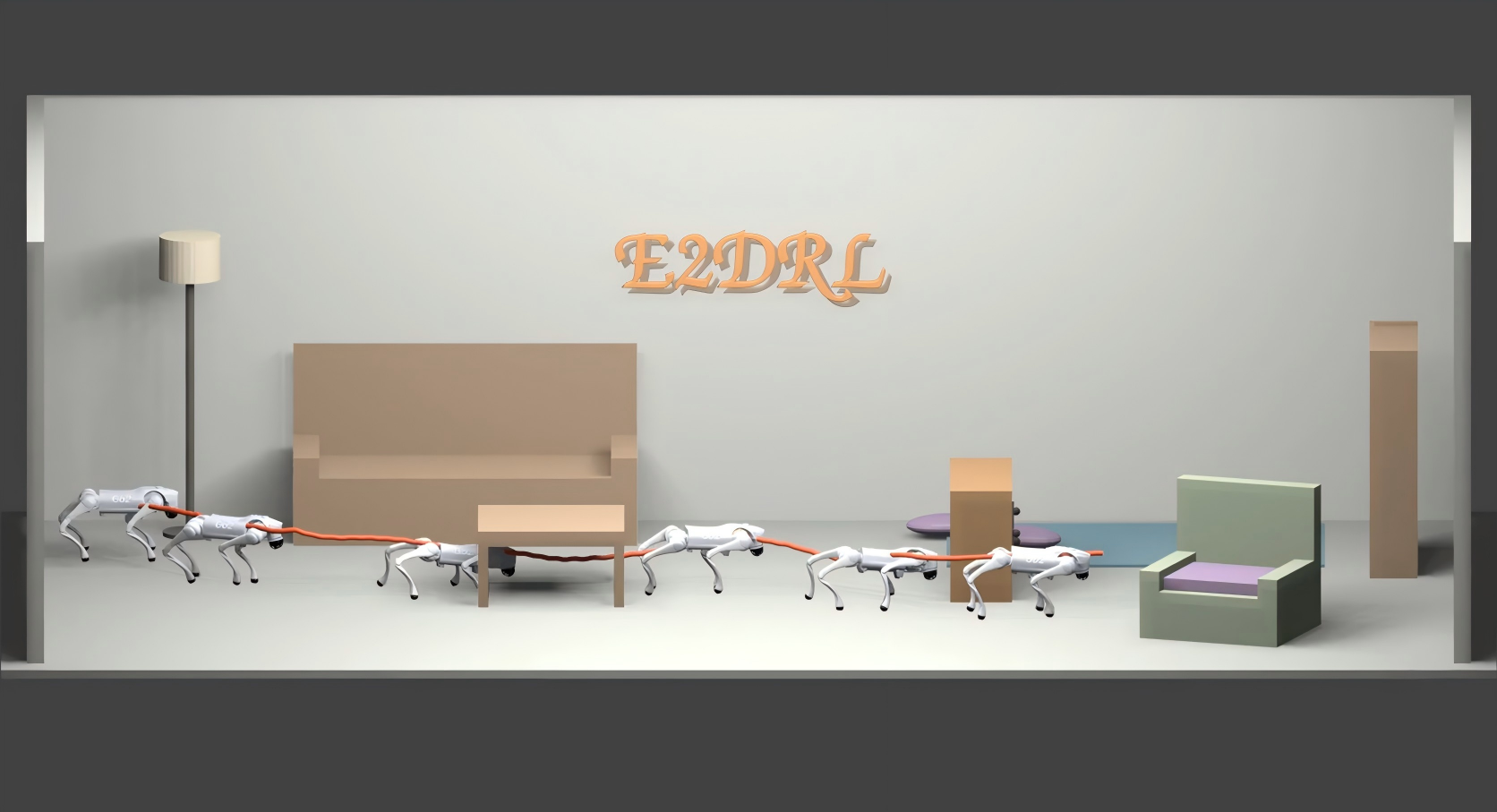}
        \label{fig:side}
    }
    \caption{3D obstacle avoidance performance of the robot in a simulated room environment.}
    \label{fig:combined_trajectory}
\end{figure}

\begin{figure}[t]
\centering
\includegraphics[width=1.0\linewidth, height=0.18\textheight]{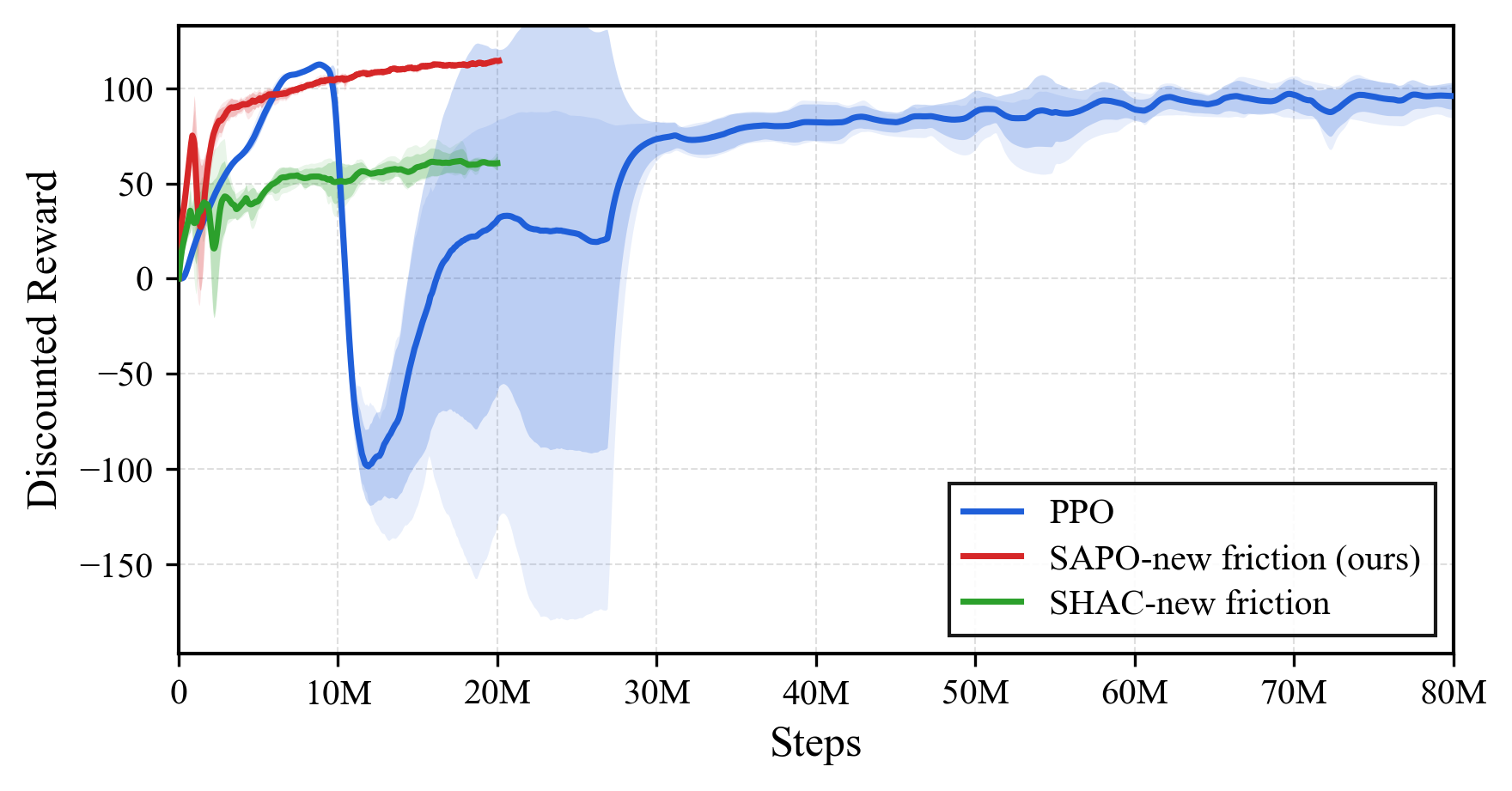}
\caption{Average discounted reward over training for PPO, SAPO, and SHAC across five random seeds. The solid line denotes the mean, the darker shaded region denotes one standard deviation, and the lighter shaded region spans the minimum and maximum across seeds.}
\label{fig:Discounted-reward}
\end{figure}

\begin{figure}[t]
    \centering
    \begin{subfigure}[t]{0.49\linewidth}
        \centering
        \includegraphics[width=\linewidth]{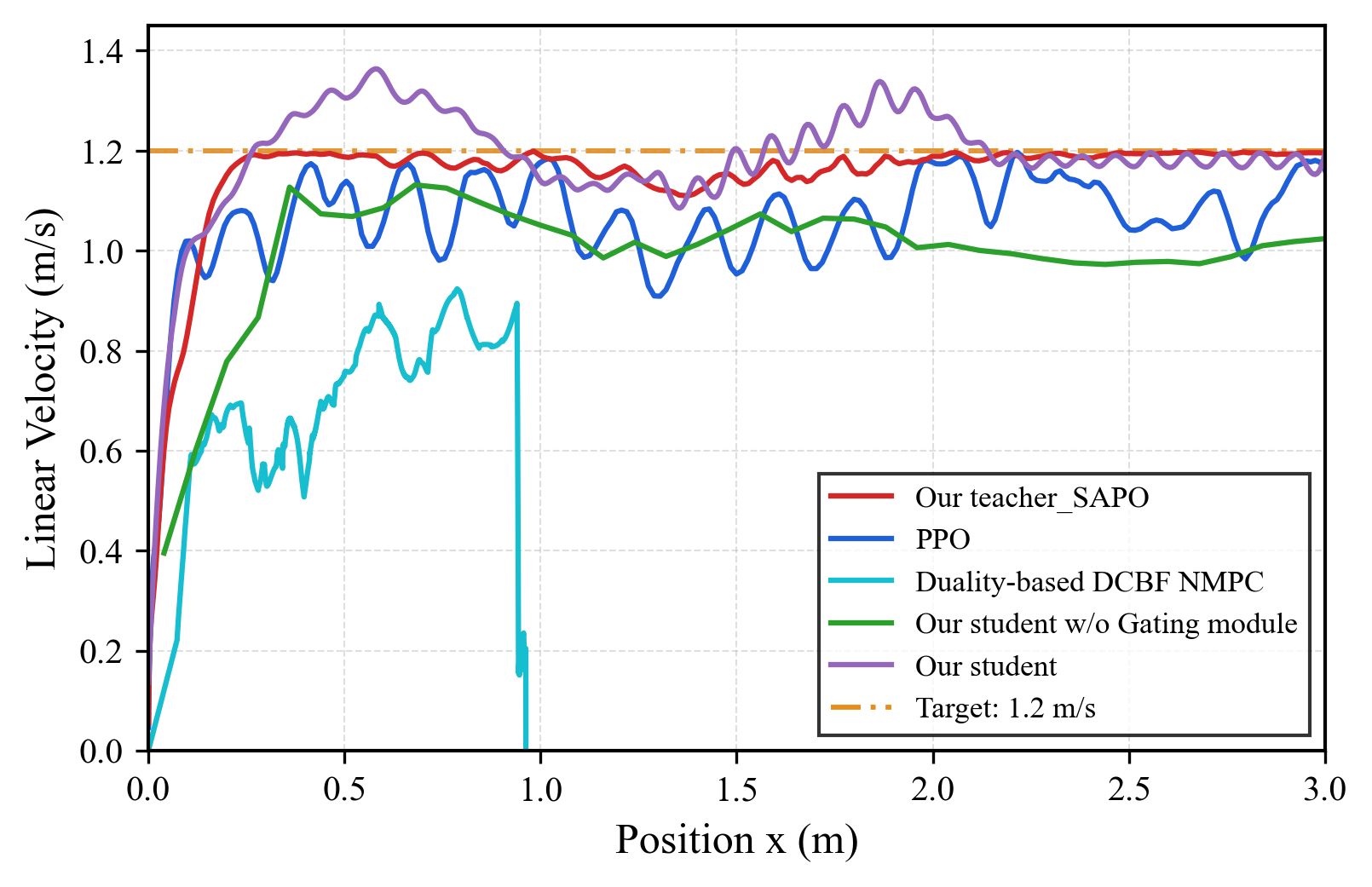}
        \caption{Overhanging obstacle.}
        \label{fig:vel-crawl}
    \end{subfigure}\hfill
    \begin{subfigure}[t]{0.49\linewidth}
        \centering
        \includegraphics[width=\linewidth]{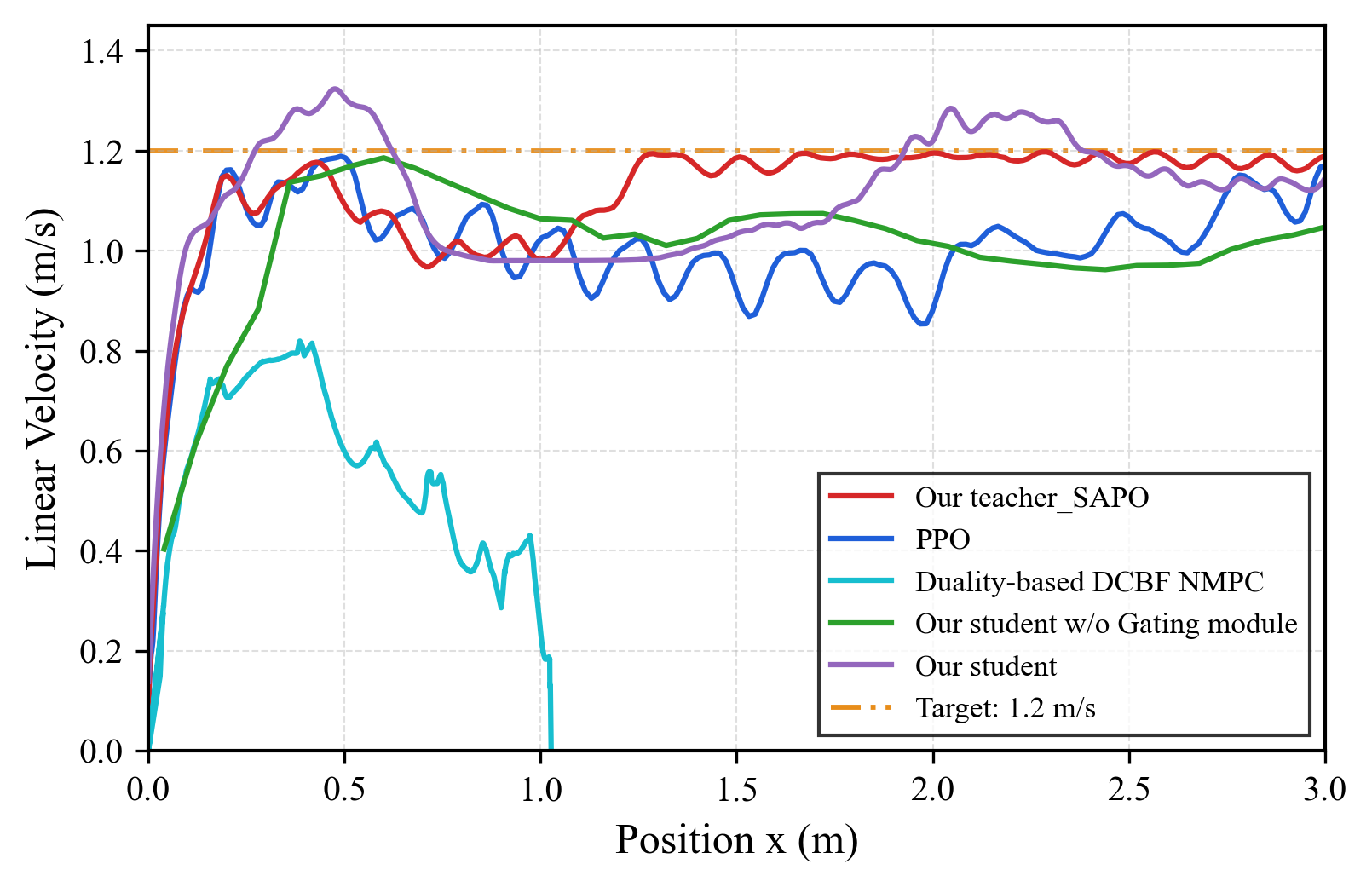}
        \caption{Ground obstacle.}
        \label{fig:vel-bypass}
    \end{subfigure}

    \caption{Average velocity-tracking performance at a commanded
    forward speed of $1.2\,$m/s in two obstacle scenarios:
    an overhanging obstacle with a lower-surface height of
    $0.28\,$m and a ground obstacle with a width of $0.8\,$m.}
    \label{fig:vel-tracking}
\end{figure}

\subsection{Simulation Experiments}
\label{sim}
\subsubsection{\textit{Simulation Setup}}

We use NVIDIA Warp \cite{macklin2022warp} as the differentiable simulator, running at a simulation frequency of 200 Hz and a control frequency of 100 Hz to limit the computation graph depth and reduce GPU memory consumption. We utilize $N = 128$ parallel environments. All training is conducted on a single NVIDIA RTX 4090 GPU. For SAPO training, we backpropagate through a 12-step differentiable rollout, with a discount factor of 0.99 and a learning rate of $2\times10^{-4}$. During training, each episode contains a single obstacle, which is randomly selected to be either an overhanging obstacle or a ground obstacle. Overhanging obstacles are sampled with clearance heights from 0.22 to 0.40~m, while ground obstacles are sampled with lateral widths from 0.60 to 1.20~m. The commanded forward velocity $V_x^{\mathrm{cmd}}$ is sampled from 1.0 to 1.5~m/s. For a fair comparison, all evaluated methods share identical task and environmental setups; in particular, the reward terms, soft obstacle formulation, and contact dynamics are identical for PPO, SHAC, and SAPO.

\begin{figure*}[t] 
    \centering

    \begin{minipage}{0.31\linewidth}
        \centering
        \scalebox{1.0}[1.0]{\includegraphics[width=1.0\linewidth]{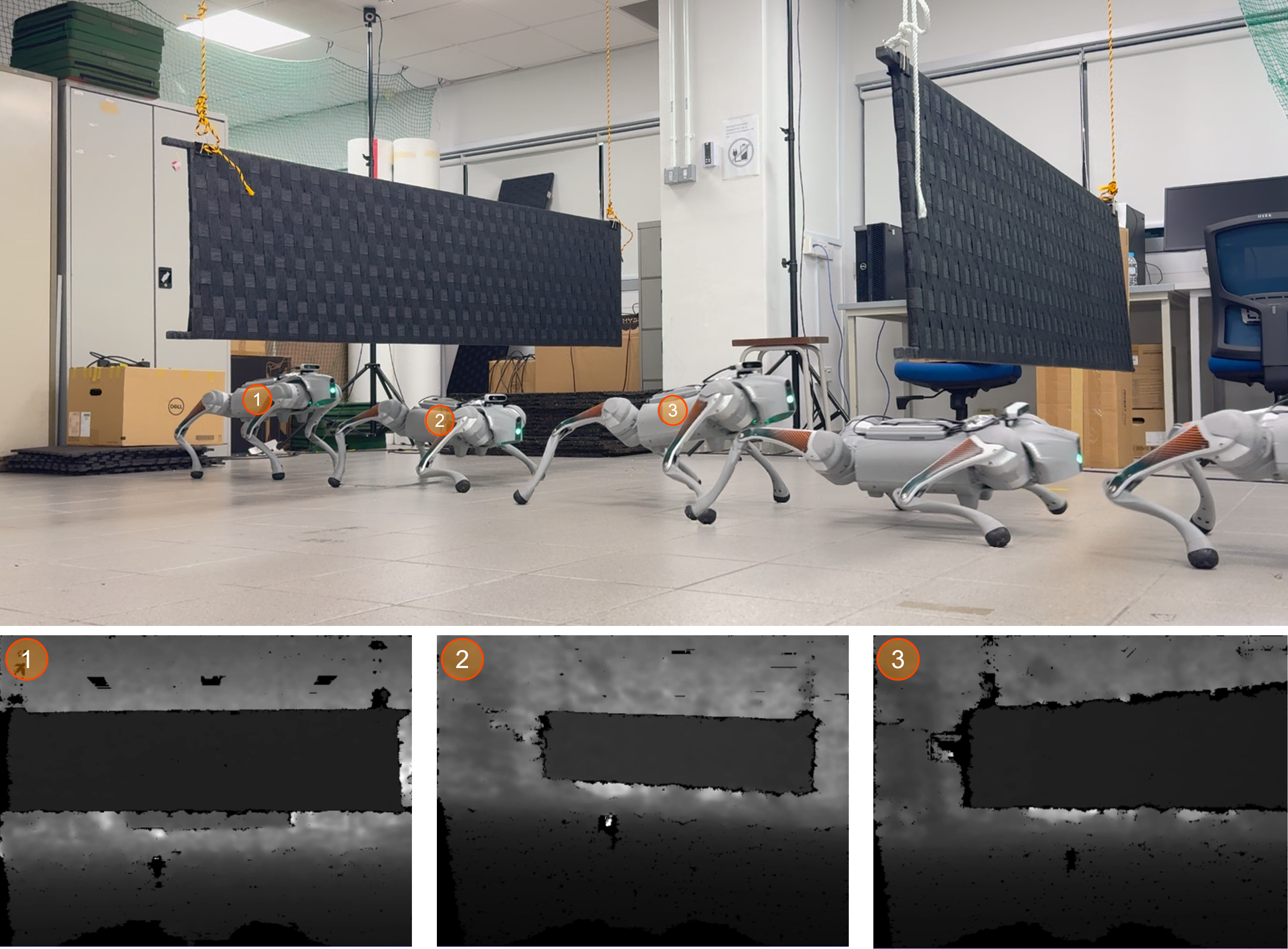}}
        \centerline{(a) Indoor overhanging obstacles}
    \end{minipage}
    \hspace{0.015\linewidth}
    \begin{minipage}{0.31\linewidth}
        \centering
        \scalebox{1.0}[1.0]{\includegraphics[width=1.0\linewidth]{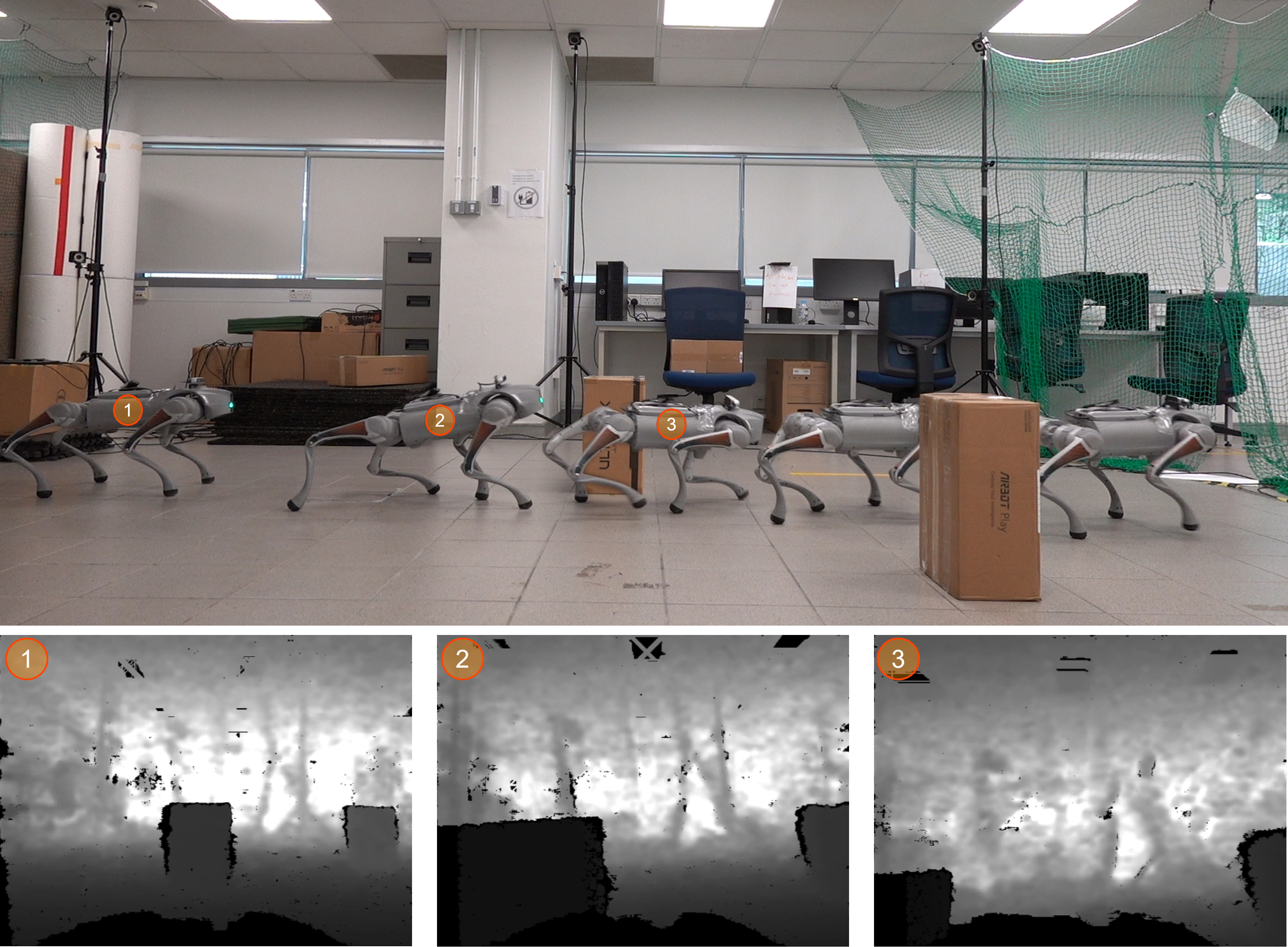}}
        \centerline{(b) Indoor ground obstacles}
    \end{minipage}
    \hspace{0.015\linewidth}
    \begin{minipage}{0.31\linewidth}
        \centering
        \scalebox{1.0}[0.993]{\includegraphics[width=1.0\linewidth]{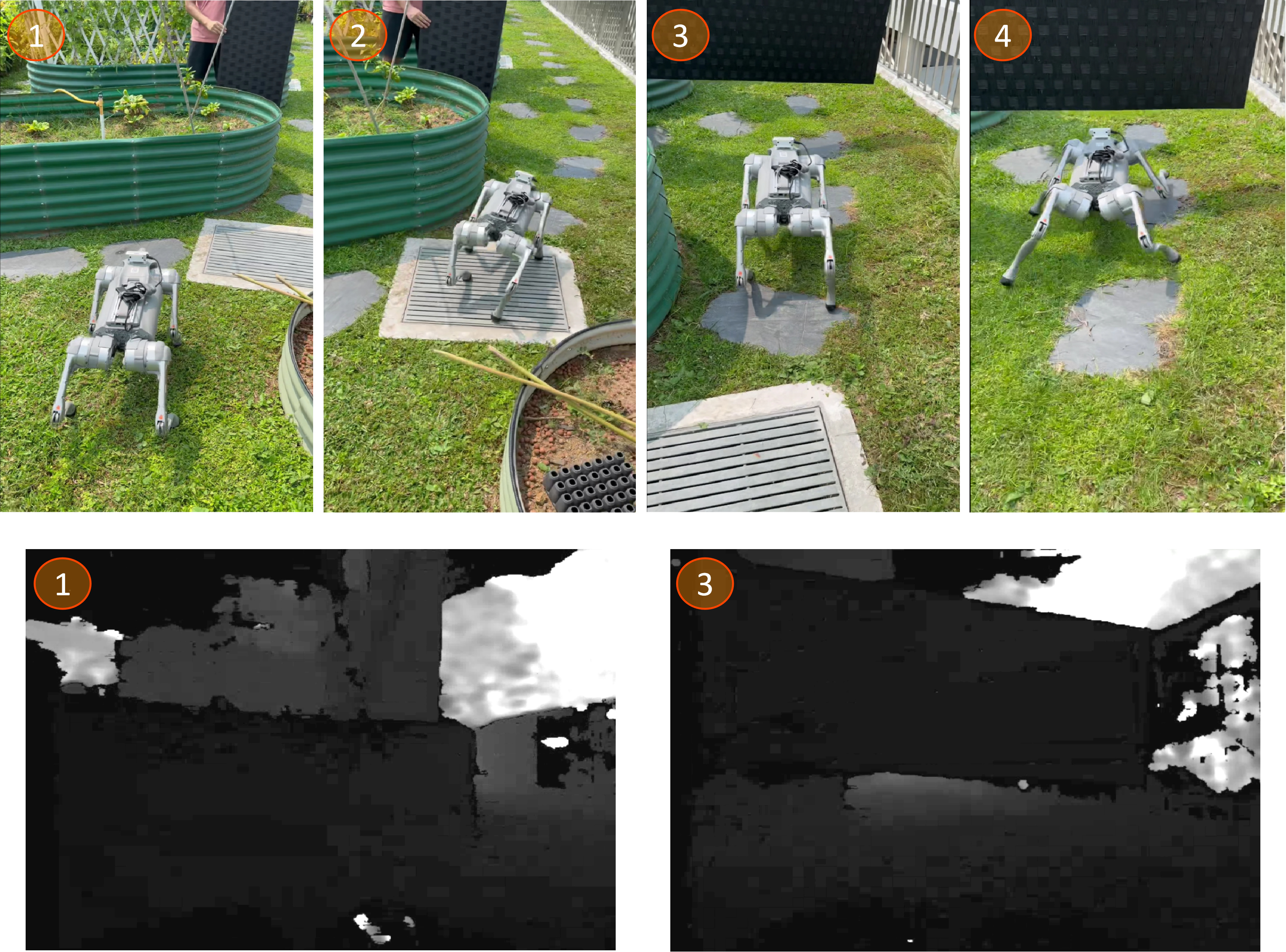}}
        \centerline{(c) Outdoor garden}
    \end{minipage}

    \vspace{0.2cm}

    \begin{minipage}{0.31\linewidth}
        \centering
        \scalebox{1.0}[1.05]{\includegraphics[width=1.0\linewidth]{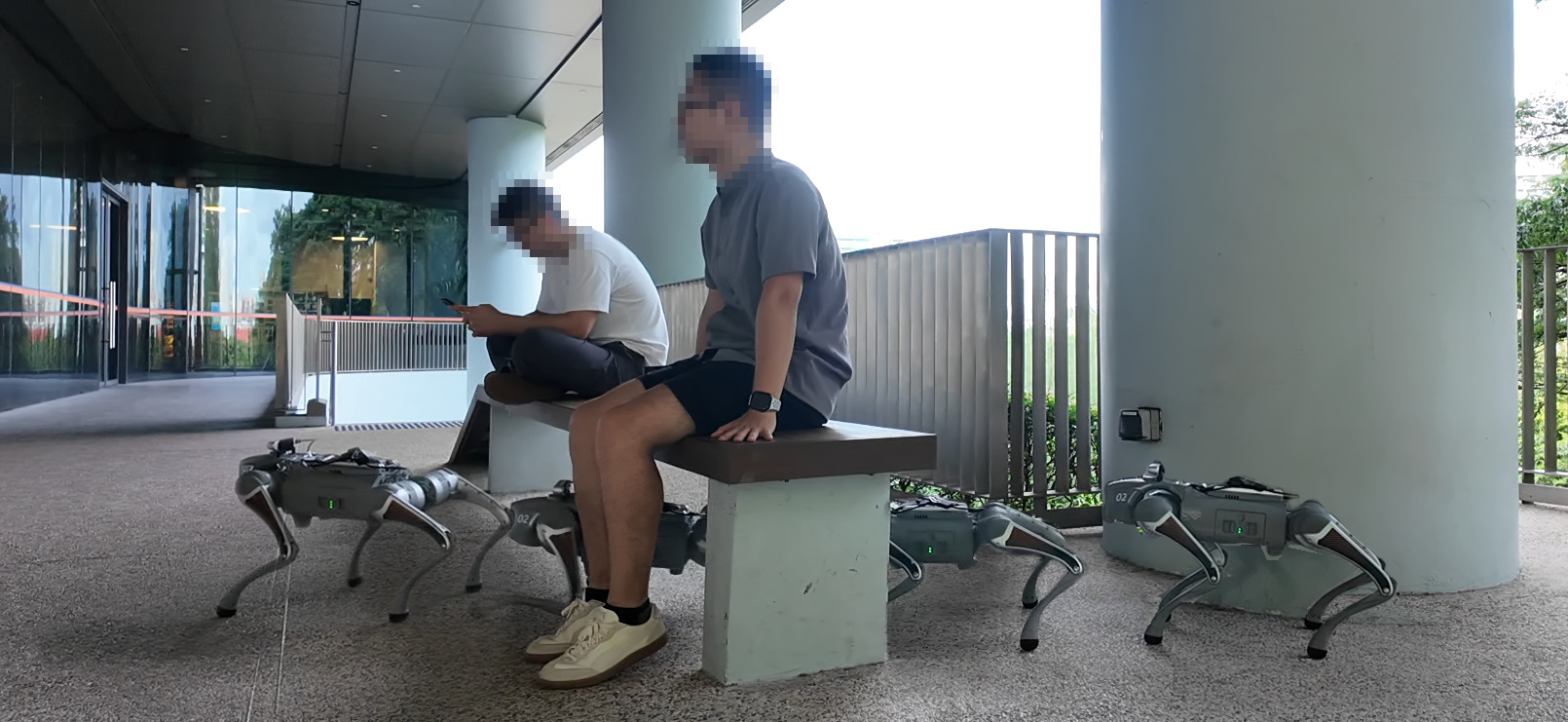}}
        \centerline{(d) Outdoor chair}
    \end{minipage}
    \hspace{0.015\linewidth}
    \begin{minipage}{0.31\linewidth}
        \centering
        \scalebox{1.0}[1.0]{\includegraphics[width=1.0\linewidth]{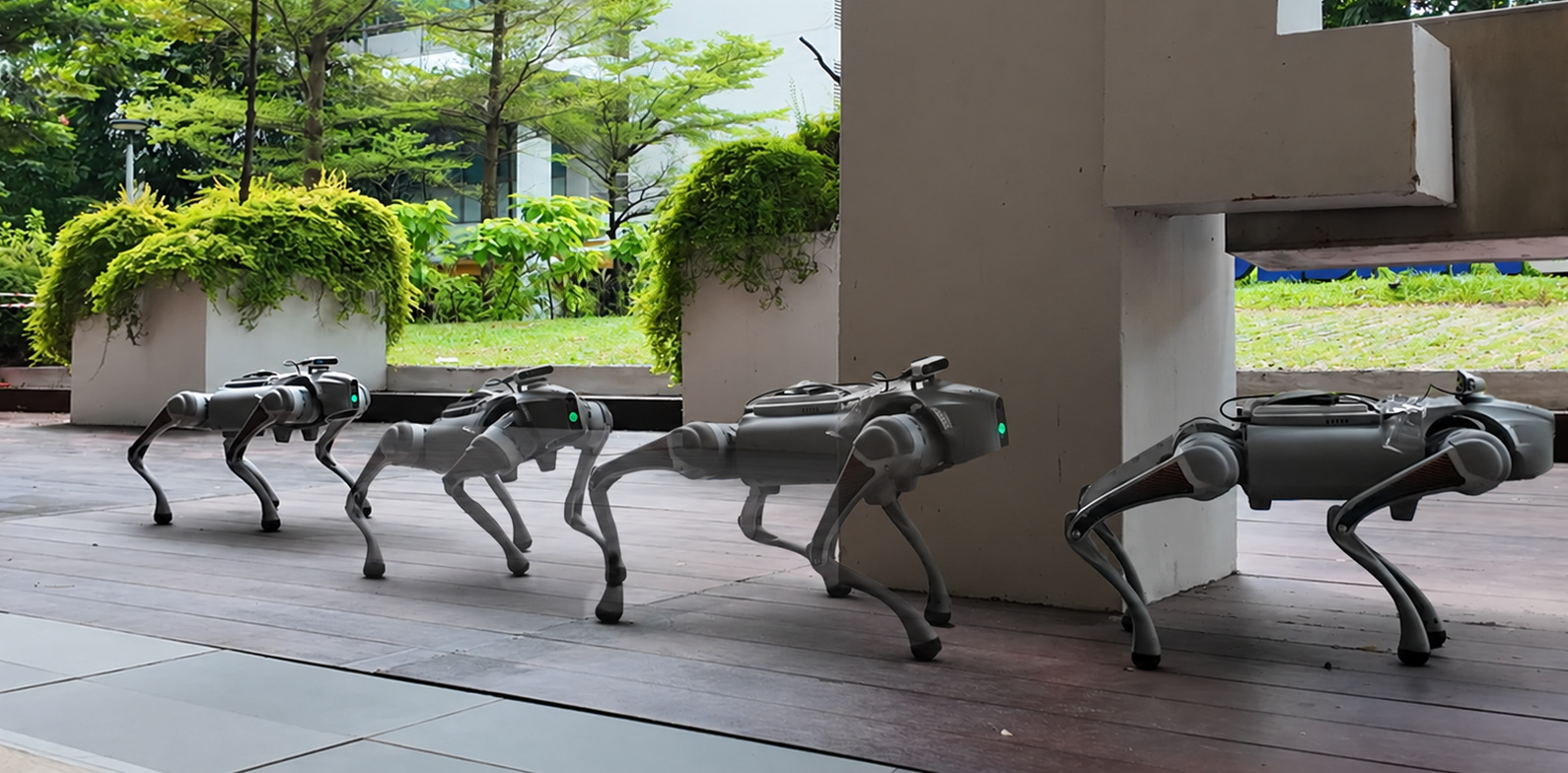}}
        \centerline{(e) Outdoor pillar}
    \end{minipage}
    \hspace{0.015\linewidth}
    \begin{minipage}{0.31\linewidth}
        \centering
        \scalebox{1.0}[1.074]{\includegraphics[width=1.0\linewidth]{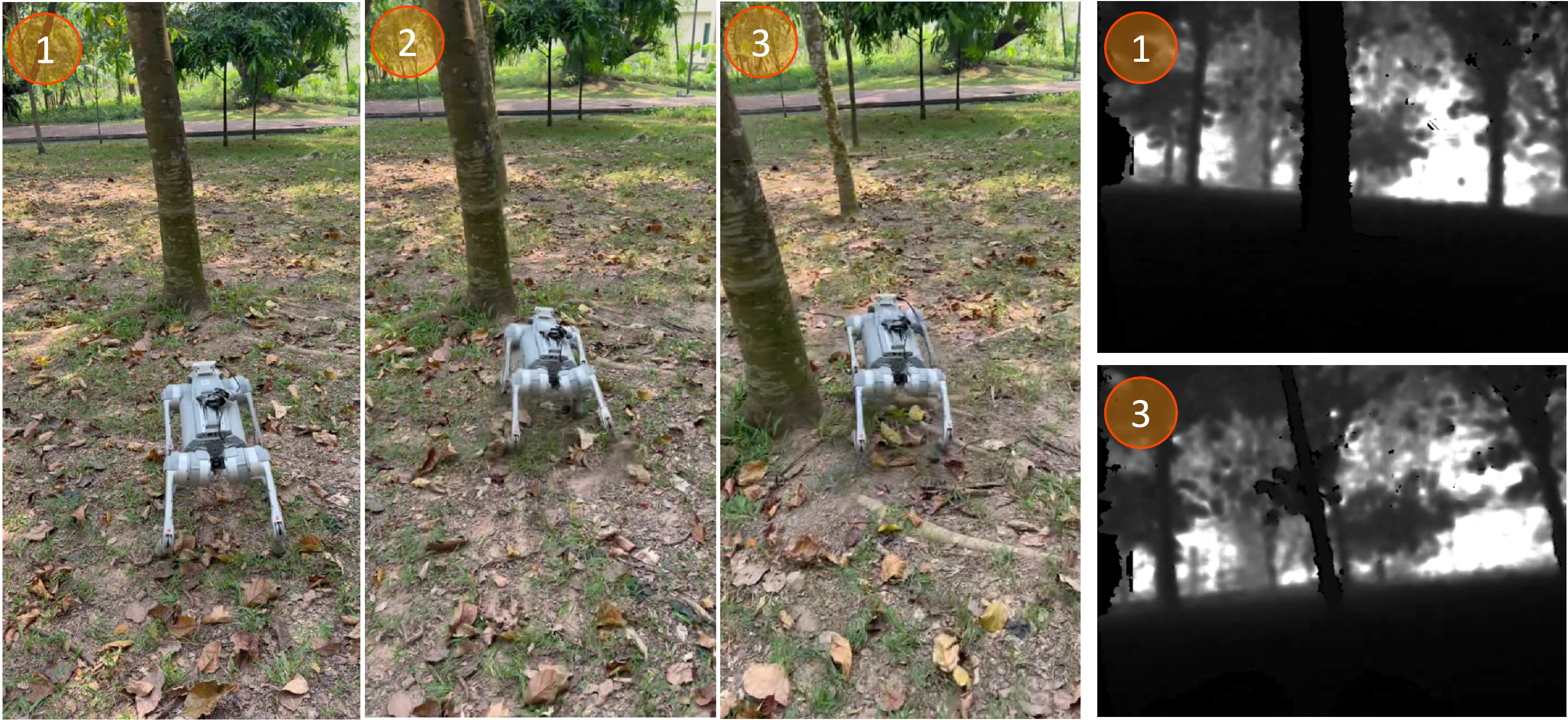}}
        \centerline{(f) Outdoor forest}
    \end{minipage}

    \caption{\textbf{Real-world static obstacle avoidance on the Unitree Go2.} The quadruped robot successfully avoids overhanging and ground obstacles across a variety of static indoor and outdoor scenarios.}
    \label{fig:static}
\end{figure*}

\begin{figure}[t]
    \centering
    \begin{subfigure}[b]{0.49\linewidth}
        \centering
        \parbox[b][0.89\linewidth][c]{\linewidth}{\includegraphics[width=\linewidth]{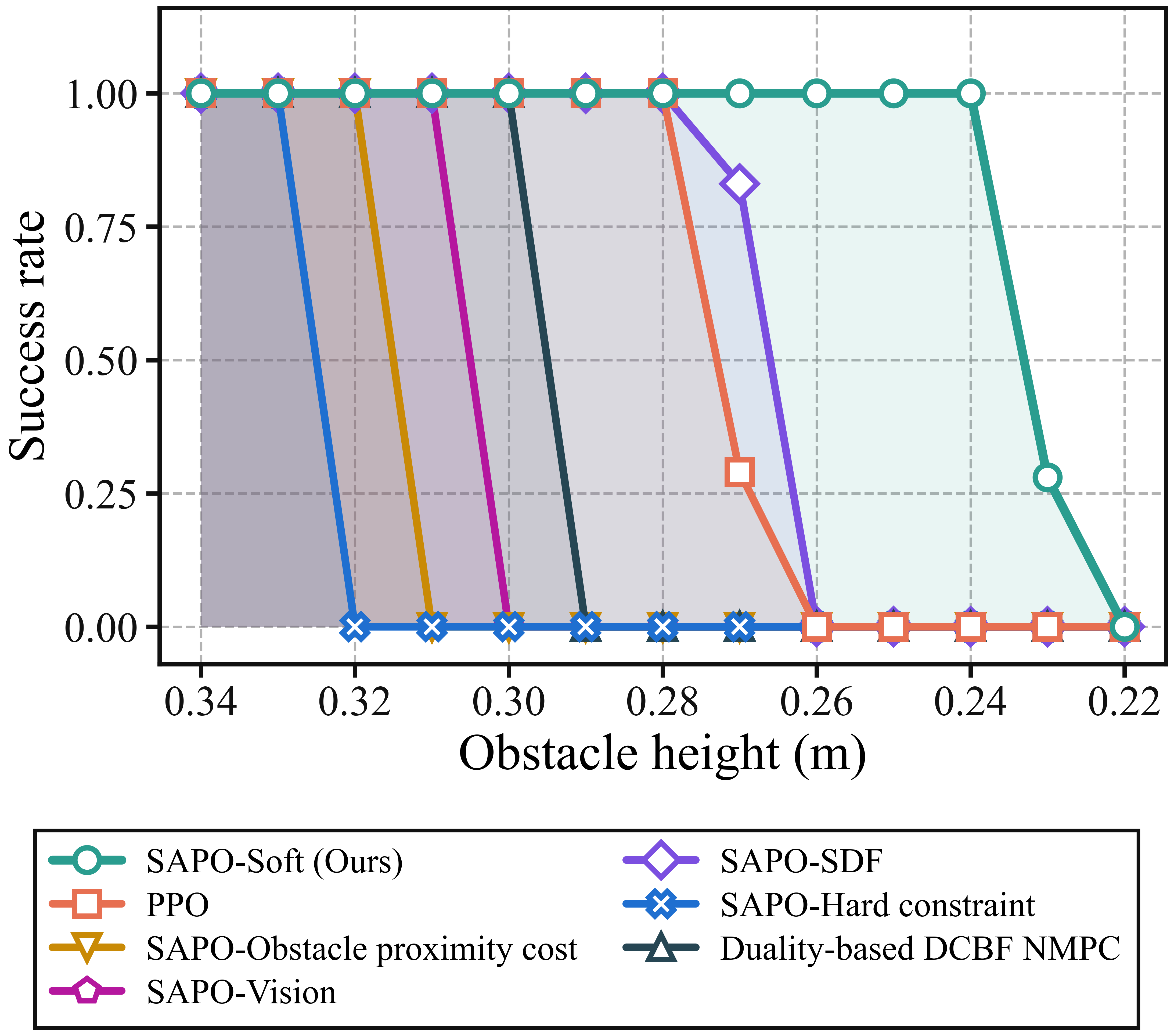}}
        \caption{Baselines (Overhanging)}
        \label{fig:success-baseline-overhanging}
    \end{subfigure}
    \hfill
    \begin{subfigure}[b]{0.49\linewidth}
        \centering
        \parbox[b][0.89\linewidth][c]{\linewidth}{\includegraphics[width=\linewidth]{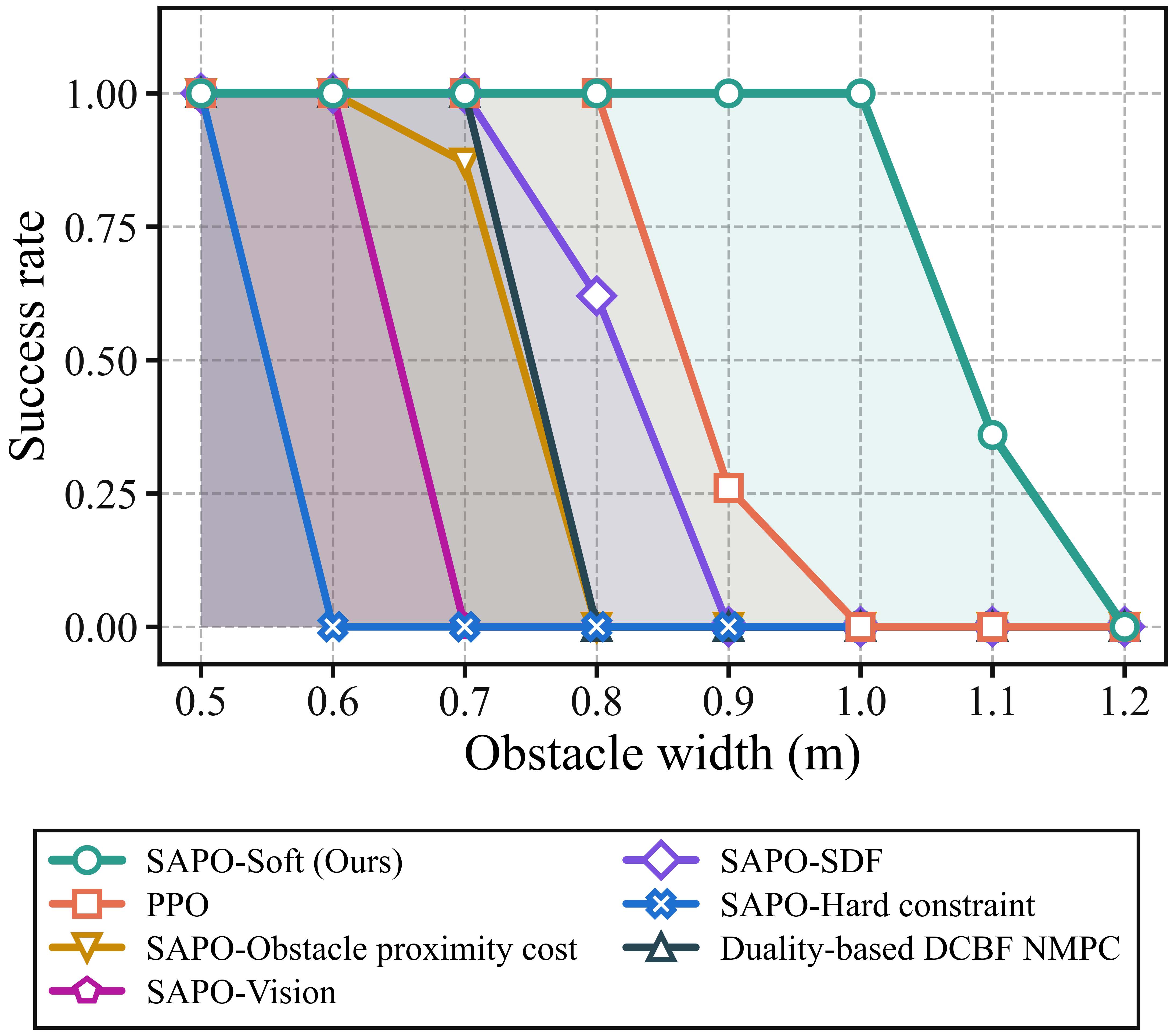}}
        \caption{Baselines (Ground)}
        \label{fig:success-baseline-ground}
    \end{subfigure}

    \vspace{0.15cm}

    \begin{subfigure}[b]{0.49\linewidth}
        \centering
        \parbox[b][0.89\linewidth][c]{\linewidth}{\includegraphics[width=\linewidth]{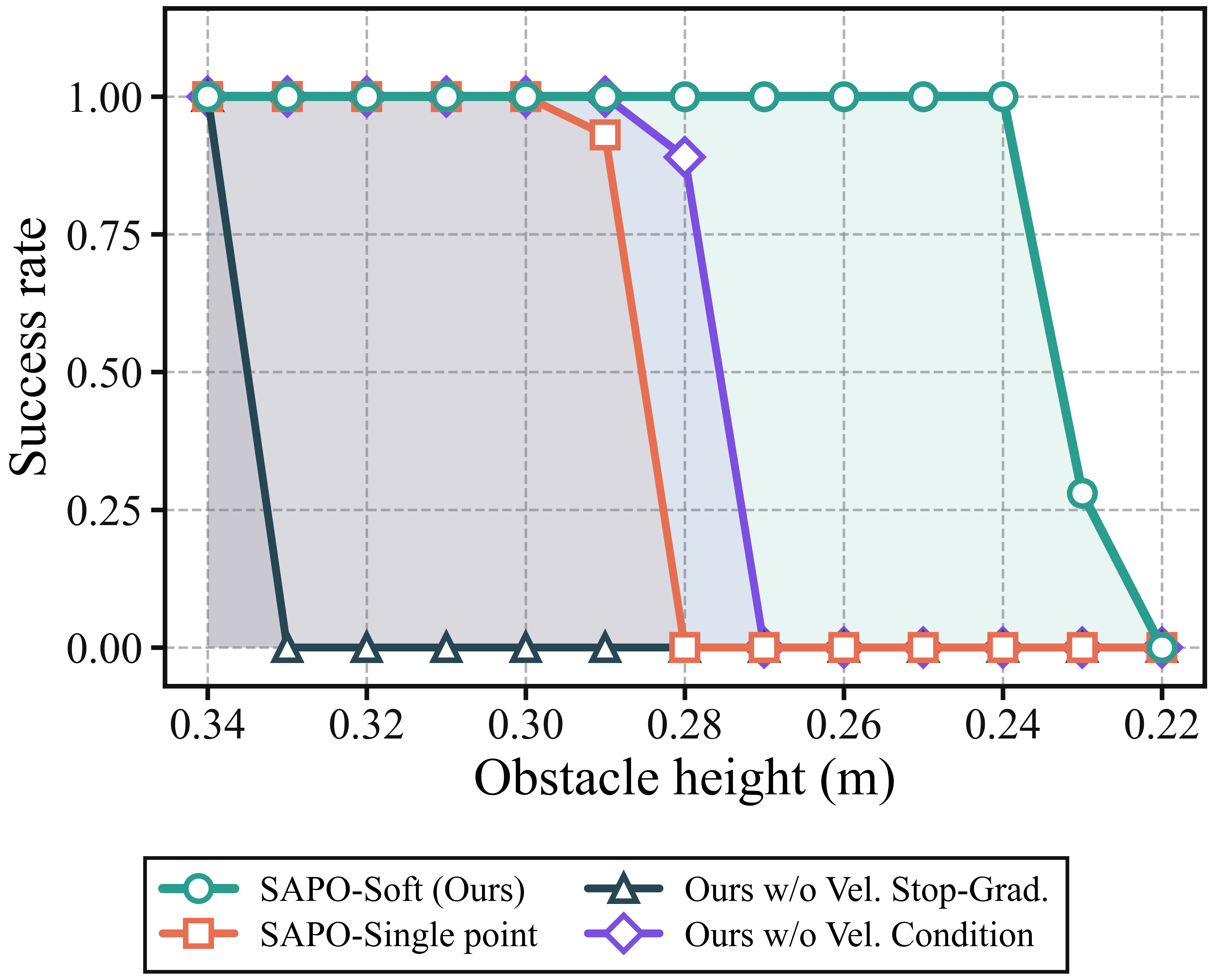}}
        \caption{Ablations (Overhanging)}
        \label{fig:success-ablation-overhanging}
    \end{subfigure}
    \hfill
    \begin{subfigure}[b]{0.49\linewidth}
        \centering
        \parbox[b][0.89\linewidth][c]{\linewidth}{\includegraphics[width=\linewidth]{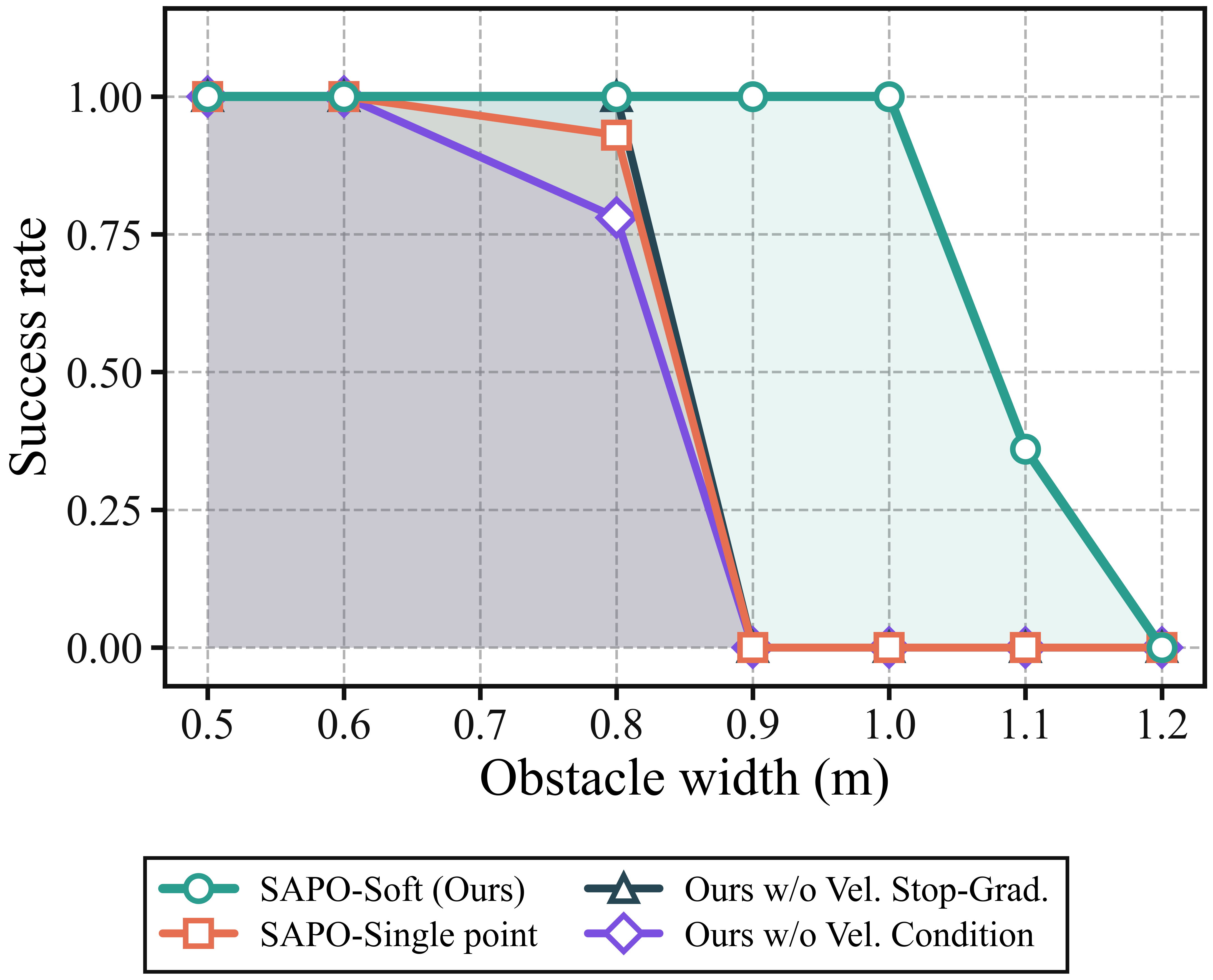}}
        \caption{Ablations (Ground)}
        \label{fig:success-ablation-ground}
    \end{subfigure}
    \caption{Collision-free success rates computed over multiple experiments under varying obstacle difficulty for overhanging and ground obstacle avoidance.}
    \label{fig:success}
\end{figure}

\subsubsection{\textit{Simulation Results}}
We first present the simulation results, followed by baseline comparisons and ablation studies. Finally, we analyze gradient stability.

\textbf{Policy Evaluation in Simulation:}
Fig.~\ref{fig:combined_trajectory} shows that the robot can avoid two distinct obstacle types while tracking the velocity commands, with the corresponding behaviors presented from both top and side views. 

    \textbf{Baseline comparison and ablation study:}
We evaluate the performance of our SAPO-based method against PPO for 3D obstacle avoidance tasks. As illustrated in Fig.~\ref{fig:Discounted-reward}, our experimental results demonstrate notable improvements in sample efficiency.
Using the avoidance success-rate plateau as the convergence criterion, SAPO converges after approximately 16M training steps, whereas PPO requires about 67M steps to reach convergence, indicating that PPO is roughly 4.2$\times$ less sample-efficient than SAPO. In addition, the final converged discounted reward achieved by SAPO is approximately 1.1 times that of PPO, demonstrating its better policy performance. Compared with SHAC, SAPO further introduces an entropy term to smooth the optimization landscape, which reduces the tendency of differentiable RL methods to get trapped in local optima. As shown in Fig.~\ref{fig:Discounted-reward}, although SAPO and SHAC exhibit comparable convergence rates, the final converged reward of SHAC is only about half that of SAPO. This substantial performance gap suggests that entropy regularization plays a critical role in helping differentiable RL escape poor local optima and achieve better solutions for contact-rich locomotion.

\begin{figure*}[t]
\centering
\includegraphics[width=1\linewidth, height=2.cm]{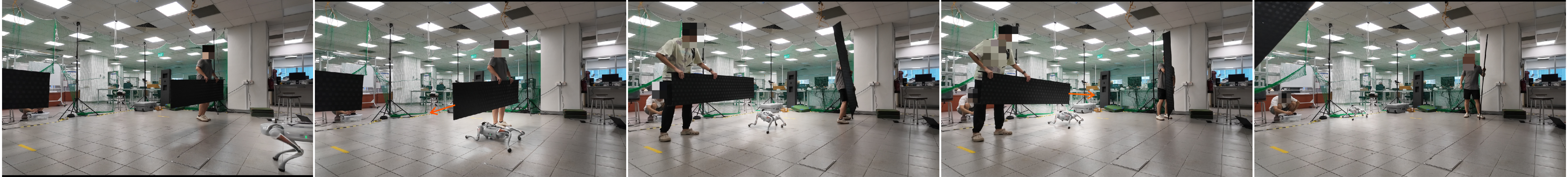}\llap{\raisebox{2pt}{\colorbox{white}{\footnotesize(a)}}\hspace{2pt}}\\[2mm]
\includegraphics[width=1\linewidth, height=2.cm]{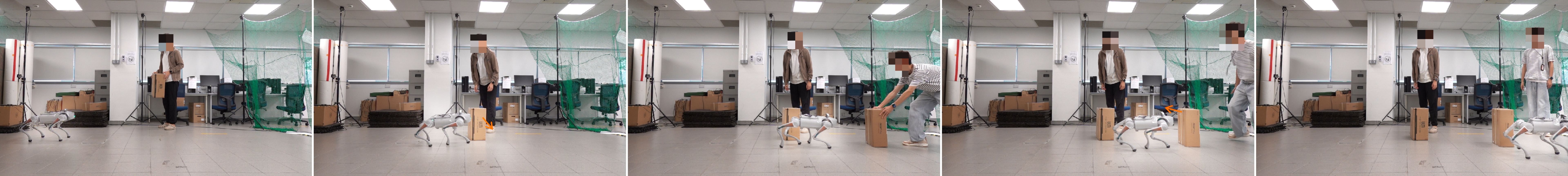}\llap{\raisebox{2pt}{\colorbox{white}{\footnotesize(b)}}\hspace{2pt}}\\[2mm]
\includegraphics[width=1\linewidth, height=2.cm]{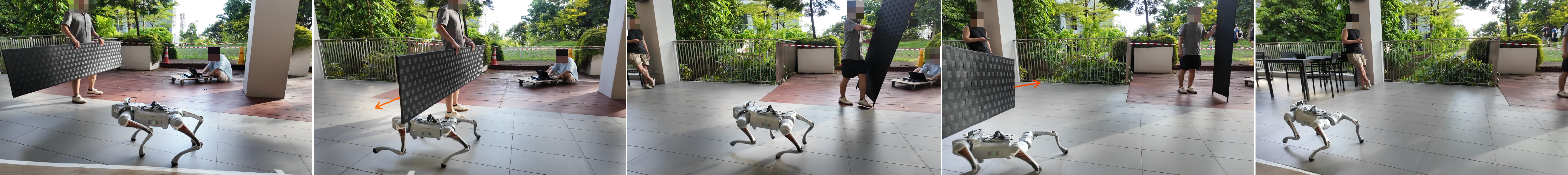}\llap{\raisebox{2pt}{\colorbox{white}{\footnotesize(c)}}\hspace{2pt}}\\[2mm]
\includegraphics[width=1\linewidth]{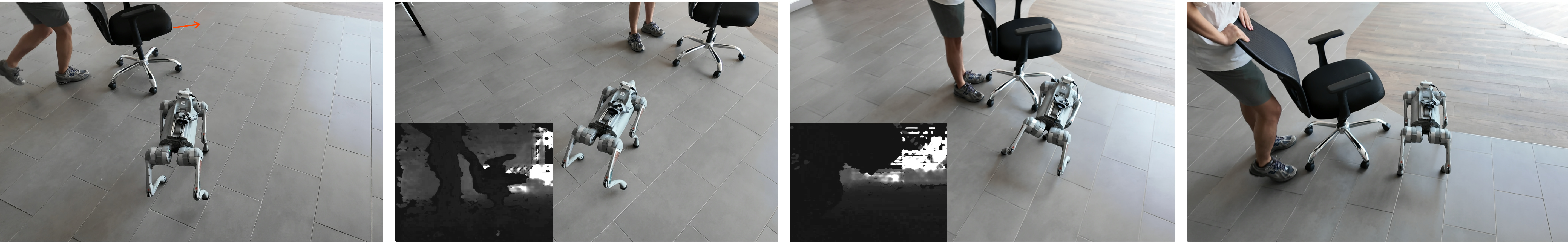}\llap{\raisebox{2pt}{\colorbox{white}{\footnotesize(d)}}\hspace{2pt}}
\includegraphics[width=\linewidth]{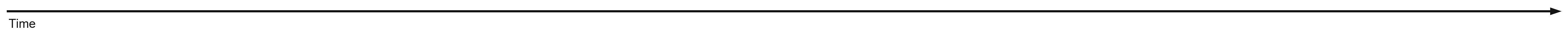}
\caption{\textbf{Real-world dynamic obstacle avoidance.} The robot reacts to sequentially introduced overhanging and ground obstacles. The images are labeled (a)–(d) from top to bottom.}
\label{fig:dynamic}
\end{figure*}

To further evaluate the learned policies over 100 environments, we examine velocity tracking in two representative 3D obstacle-avoidance scenarios with a target velocity of 1.2~m/s: passing beneath an overhanging obstacle with a clearance height of 0.28~m and bypassing a ground obstacle with a width of 0.8~m. We compare our approach against PPO and the tuned Duality-based DCBF NMPC baseline described above. As illustrated in Fig.~\ref{fig:vel-tracking}, our method achieves faster convergence to the commanded velocity than the NMPC baseline, while maintaining precise tracking throughout the tasks with only minor deviations during obstacle avoidance. In contrast, PPO suffers from clear oscillations and tracking errors, and the NMPC approach ultimately fails to complete the traversal and comes to a halt. These results demonstrate that our differentiable RL formulation achieves an excellent balance between velocity tracking and obstacle avoidance. Fig.~\ref{fig:vel-tracking} also compares the student policy with and without the gating module. In obstacle-free segments, the gated policy better tracks the $1.2\,$m/s command than the ungated policy, indicating that the gating module improves nominal velocity-tracking performance.

Fig.~\ref{fig:success} reports the success rate averaged over 100 environments for each combination of commanded velocity, sampled in 0.1~m/s increments, and obstacle geometry, swept in 0.01~m and 0.1~m increments for overhanging and ground obstacles, respectively. Our method passes beneath overhanging obstacles with clearance heights down to 0.24~m, close to the minimum body height of the Unitree Go2, and bypasses ground obstacles with widths up to 1.0~m. In contrast, PPO and the tuned Duality-based DCBF NMPC baseline fail at less challenging configurations. The obstacle-proximity baseline achieves lower success rates than our method for both obstacle types, as its proximity reward discourages the close boundary-following needed for tight geometries. The SDF baseline also achieves lower success rates, since its reward can favor high-speed traversal with brief, shallow penetration. The distilled student policy handles more challenging obstacles than the policy trained directly with visual observations.

As shown in Fig.~\ref{fig:success}, we conduct ablation studies to validate the proposed obstacle-interaction design, keeping all other training and evaluation configurations identical. First, replacing the soft penetration penalty with hard termination upon contact significantly reduces the feasible obstacle range: our method achieves height and width limits of 0.24~m and 1.0~m for overhanging and ground obstacles, whereas the hard termination baseline degrades to 0.33~m and 0.58~m, respectively. Unlike hard termination which yields discontinuity, our soft penalty provides dense gradients that guide body-lowering and lateral avoidance. Second, removing the velocity-conditioned stop-gradient shaping leads to increased collision failures, restricting the feasible height and width to only 0.34~m and 0.85~m for overhanging and ground obstacles, respectively. Without the stop-gradient, the optimizer tends to reduce penalties by suppressing forward velocity rather than learning obstacle avoidance. In addition, removing velocity conditioning reduces success rates on both tasks, as the robot becomes more prone to high-speed penetration. Finally, using only a single keypoint on the base reduces success rates on both tasks because it provides incomplete feedback for whole-body posture changes.

\begin{figure}[t]
    \centering
    \begin{subfigure}[t]{0.49\linewidth}
        \centering
        \includegraphics[width=\linewidth]{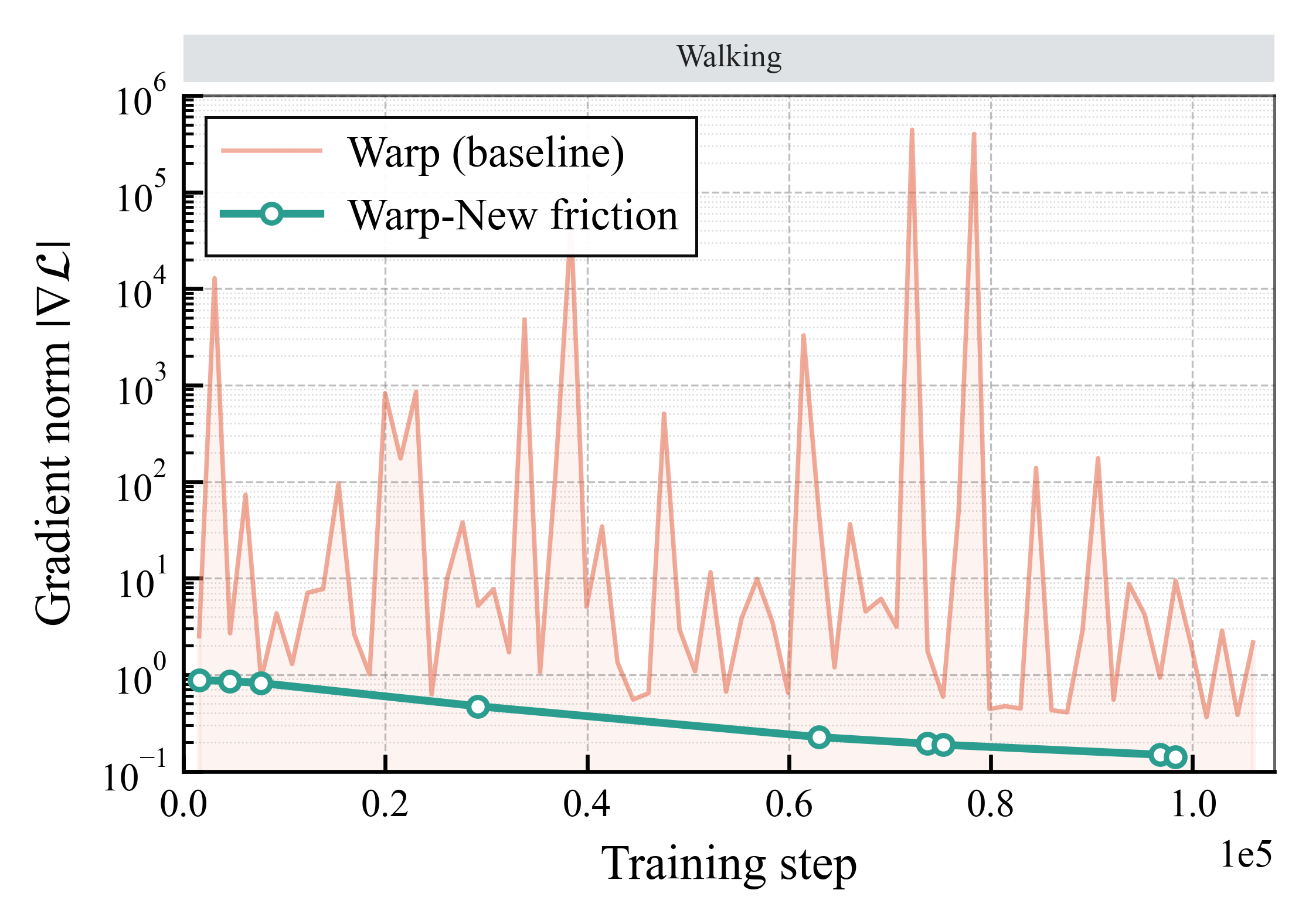}
        \caption{Walking stage.}
        \label{fig:gradient-walking}
    \end{subfigure}\hfill
    \begin{subfigure}[t]{0.49\linewidth}
        \centering
        \includegraphics[width=\linewidth]{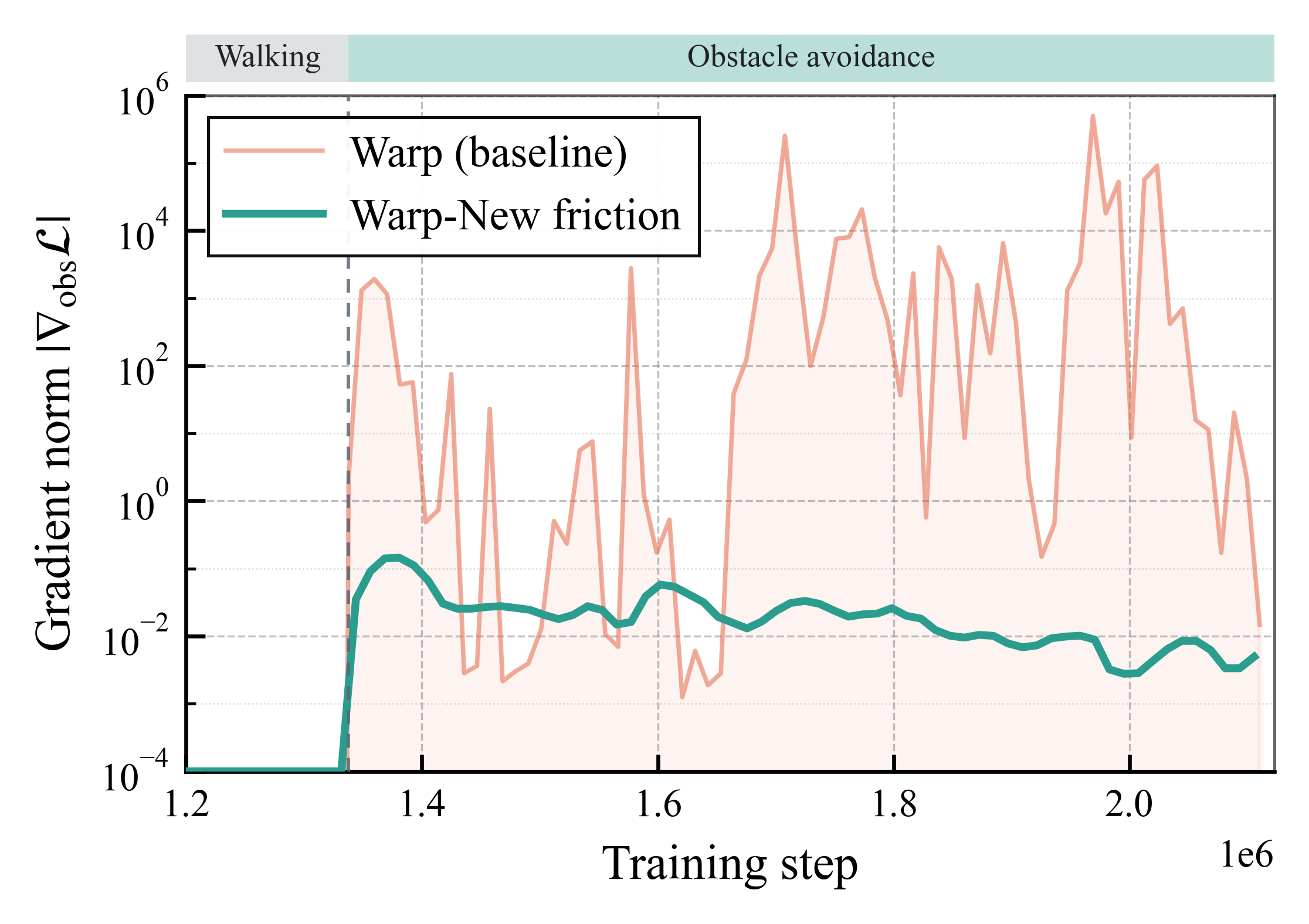}
        \caption{Obstacle-avoidance stage.}
        \label{fig:gradient-obstacle}
    \end{subfigure}

    \caption{Actor-gradient magnitude during SAPO training with the
    differentiable friction model in \eqref{eq:fc}--\eqref{eq:ft} and the
    default Warp friction model: (a) total actor gradient
    $|\nabla\mathcal{L}|$ during the walking stage, and (b) obstacle-term
    gradient $|\nabla_{\mathrm{obs}}\mathcal{L}|$ across the transition into
    the obstacle-avoidance stage.}
    \label{fig:gradient}
\end{figure}

\textbf{Gradient stability:}
As shown in Fig.~\ref{fig:gradient}, the contact model in \eqref{eq:fc}--\eqref{eq:ft} keeps the reported actor-gradient magnitude below 1.0, avoiding gradient spikes during foot strikes. Specifically, Fig.~\ref{fig:gradient}(a) shows the gradient norm of the entire actor, while Fig.~\ref{fig:gradient}(b) shows the gradient norm induced by the soft differentiable obstacle-interaction formulation.

Overall, these results validate the effectiveness of our method in challenging 3D obstacle-avoidance scenarios.

\subsection{\textit{Real-World Experiments}}
\label{real}
\subsubsection{\textit{Real-World Experimental Setup}}

We use a Unitree Go2 equipped with an NVIDIA Jetson Orin NX for onboard deployment and an Intel RealSense D435 for onboard visual perception.
We conducted physical experiments for both static and dynamic 3D obstacles to validate our approach's effectiveness in real-world scenarios. The experiments were also conducted indoors and outdoors to ensure the generalization of the policy.  

\subsubsection{Results in Static Environments}

We evaluate the robot in static obstacle environments, as shown in Fig.~\ref{fig:static}. In the indoor overhanging scenario depicted in Fig.~\ref{fig:static}(a), the robot sequentially crawls beneath two obstacles with decreasing clearances of 0.38~m and 0.27~m. For indoor ground obstacles, illustrated in Fig.~\ref{fig:static}(b), it executes smooth lateral avoidance while maintaining forward locomotion. The policy effectively extracts obstacle geometry from depth data, suppresses background clutter, and handles consecutive interactions. Across 5 trials per scenario, the robot achieved success rates of 5/5 for ground obstacles and 4/5 for overhanging obstacles, with the failure due to a collision under the lower clearance. The performance degradation can be attributed in part to policy distillation and the sim-to-real gap in depth images. Furthermore, the policy generalizes to diverse outdoor conditions. In Fig.~\ref{fig:static}(c), the robot consecutively avoids a ground obstacle and an overhanging obstacle on grassy terrain. As demonstrated in Fig.~\ref{fig:static}(d) and Fig.~\ref{fig:static}(e), the robot successfully crawls under a chair and bypasses pillars in two different outdoor scenes, respectively. As shown in Fig.~\ref{fig:static}(f), the robot bypasses trees while traversing uneven terrain in a forest.

\subsubsection{Results in Dynamic Environments}
We also evaluate the proposed approach in dynamic environments with time-varying obstacle layouts. Representative results are shown in Fig.~\ref{fig:dynamic}. In the dynamic overhanging obstacle case shown in Fig.~\ref{fig:dynamic}(a), the robot successfully passes beneath two sequentially introduced overhanging obstacles during locomotion. 
In the dynamic ground obstacle scenario in Fig.~\ref{fig:dynamic}(b), the robot avoids dynamically introduced ground obstacles during forward motion. 
Additional outdoor dynamic experiments are shown in Figs.~\ref{fig:dynamic}(c)--(d). Fig.~\ref{fig:dynamic}(c) demonstrates that the robot can consecutively avoid multiple dynamic overhanging obstacles, while Fig.~\ref{fig:dynamic}(d) shows that it can avoid an irregularly shaped chair that suddenly appears in its path. Based on the previously trained policy, about 2 h of additional training enables the policy to handle such irregular geometries, demonstrating the extensibility of our method to more complex obstacles.

Overall, these real-world trials show that the proposed differentiable RL framework effectively enables continuous 3D obstacle avoidance. The learned policy generalizes across indoor and outdoor scenes and handles both static and dynamic obstacles.

\section{CONCLUSIONS}

This paper presents E2DRL, an end-to-end differentiable reinforcement learning framework for 3D obstacle avoidance in quadruped robots that bridges first-order differentiable reinforcement learning and real-world deployment. Unlike prior differentiable RL approaches for legged robots that are restricted to blind walking or simulated 2D tasks, our method enables a quadruped robot to negotiate both overhanging obstacles and ground obstacles in a unified framework, while supporting sim-to-real transfer to physical hardware.

To achieve this, we propose a training pipeline that first optimizes a privileged teacher policy in a differentiable simulator via SAPO. Rather than relying on hard collision terminations, our soft obstacle-interaction formulation employs continuous multi-keypoint penetration penalties and velocity-conditioned stop-gradient shaping. This provides stable gradient signals that enable the robot to crawl beneath overhanging obstacles and bypass ground obstacles. For real-world deployment, this learned behavior is distilled into a recurrent depth-based student policy via DAgger and fused with a proprioceptive base policy through a terrain-aware gating module. In simulation, E2DRL demonstrates a $4.2\times$ improvement in sample efficiency over PPO according to the avoidance success-rate plateau criterion, alongside stronger velocity tracking and obstacle avoidance capabilities. Finally, hardware experiments on a Unitree Go2 validate reliable sim-to-real transfer across indoor and outdoor scenarios involving static and dynamically introduced obstacles.

Although E2DRL demonstrates effective sim-to-real transfer, its performance still depends on the quality of onboard depth perception and may degrade under reflective, transparent, or highly noisy sensing conditions. Future work will focus on improving perceptual robustness and extending the framework toward more complex scenarios.

\section*{Acknowledgments}
This work was supported by the MOE AcRF Tier 1
24-1234-P0001 Funding, NUS Presidential Young
Professorship, and NVIDIA Academic Grant.

\FloatBarrier

\bibliographystyle{IEEEtran}
\bibliography{ref}

\endgroup

\clearpage
\onecolumn
\begingroup
\raggedbottom
\setlength{\intextsep}{6pt plus 1pt minus 1pt}
\setcounter{secnumdepth}{2}
\setcounter{subsection}{0}
\setcounter{figure}{0}
\setcounter{table}{0}
\setcounter{equation}{0}
\renewcommand{\thesubsection}{S\arabic{subsection}}
\renewcommand{\thesubsectiondis}{\thesubsection.}
\renewcommand{\thefigure}{S\arabic{figure}}
\renewcommand{\thetable}{S\arabic{table}}
\renewcommand{\theequation}{S\arabic{equation}}
\renewcommand{\theHsubsection}{supplement.\arabic{subsection}}
\renewcommand{\theHfigure}{supplement.\arabic{figure}}
\renewcommand{\theHtable}{supplement.\arabic{table}}
\renewcommand{\theHequation}{supplement.\arabic{equation}}
\phantomsection
\section*{Supplementary Material}
\addcontentsline{toc}{section}{Supplementary Material}
\begin{center}
\large\textit{E2DRL: End-to-End Differentiable Reinforcement Learning for 3D Obstacle Avoidance in Quadruped Robots}
\end{center}

This supplement provides implementation details, stop-gradient analysis, teacher--student transfer results, training-time comparisons, and parameter sensitivity studies for E2DRL.

\subsection{Network Architecture and Deployment}
\label{sm:architecture}

\noindent\textbf{Privileged teacher and visual student.}
The teacher is an MLP with hidden layers of $[256,128,64]$. The student encodes depth observations using three convolutional layers with $[16,32,32]$ channels. The flattened visual features are concatenated with the proprioceptive state and passed to a GRU with a hidden dimension of 256. An MLP with hidden layers of $[256,128,64]$ maps the GRU output to 12 joint-position targets. The GRU hidden state is reset at the beginning of each episode to prevent cross-episode information leakage.

\medskip
\noindent\textbf{Gating and deployment.}
The learned gating module described in Section~\ref{sec:deployment} estimates obstacle proximity rather than taking the minimum raw depth: the nearest return typically comes from the ground at approximately $0.5\,$m, already below the $0.6\,$m activation threshold. Explicit ground segmentation would depend on camera pose, calibration, terrain geometry, and sensor noise. The student's mapping from depth and proprioception to joint-position targets is end-to-end; the teacher, student, nominal policy, and gating module are trained separately.

\subsection{Training and Evaluation Protocol}
\label{sm:protocol}

\noindent\textbf{Shared training settings.}
The PPO, SHAC, and SAPO comparisons share actor and critic architectures, observation and action spaces, rewards, obstacle and contact formulations, and curriculum, with the simulation setup specified in Section~\ref{sim}.

All methods begin with a walking stage. Training switches to obstacle avoidance when the policy achieves a 70\% success rate in reaching a target $4.5\,$m away. Obstacle configurations are subsequently randomized within a difficulty range whose upper bound increases after five consecutive collision-free traversals. An episode is limited to 3000 simulation steps and terminates early if the center-of-mass height falls below $0.12\,$m, the absolute roll angle exceeds $0.5$, or the absolute pitch angle exceeds $0.5$.

\medskip
\noindent\textbf{Training variability.}
PPO, SHAC, and SAPO are each trained with five random seeds: 800, 900, 1000, 1100, and 1200. Figure~\ref{fig:Discounted-reward} in the main paper reports the mean, one standard deviation, and the minimum-to-maximum range across these seeds. The shaded bands represent empirical variability, not confidence intervals.

\medskip
\noindent\textbf{Obstacle-limit criterion.}
For the parameter sweeps in Section~\ref{sm:parameters}, the reported obstacle limit is the most challenging geometry at which the success rate is at least 90\%. A smaller minimum overhanging clearance and a larger maximum ground-obstacle width indicate better avoidance capability. These simulation limits are distinct from the hardware trial results in Section~\ref{real}.

\subsection{Body Keypoints and Baseline Implementation}
\label{sm:interaction}

\noindent\textbf{Body keypoints and penetration geometry.}
The obstacle-interaction reward in \eqref{eq:penetration_reward} uses 19 keypoints distributed over the robot body. Figure~\ref{fig:sm_keypoints} illustrates their placement and penetration depths. The single-keypoint ablation uses the center of the top surface of the robot's base.

\begin{figure}[H]
    \centering
    \includegraphics[width=0.45\linewidth,keepaspectratio]{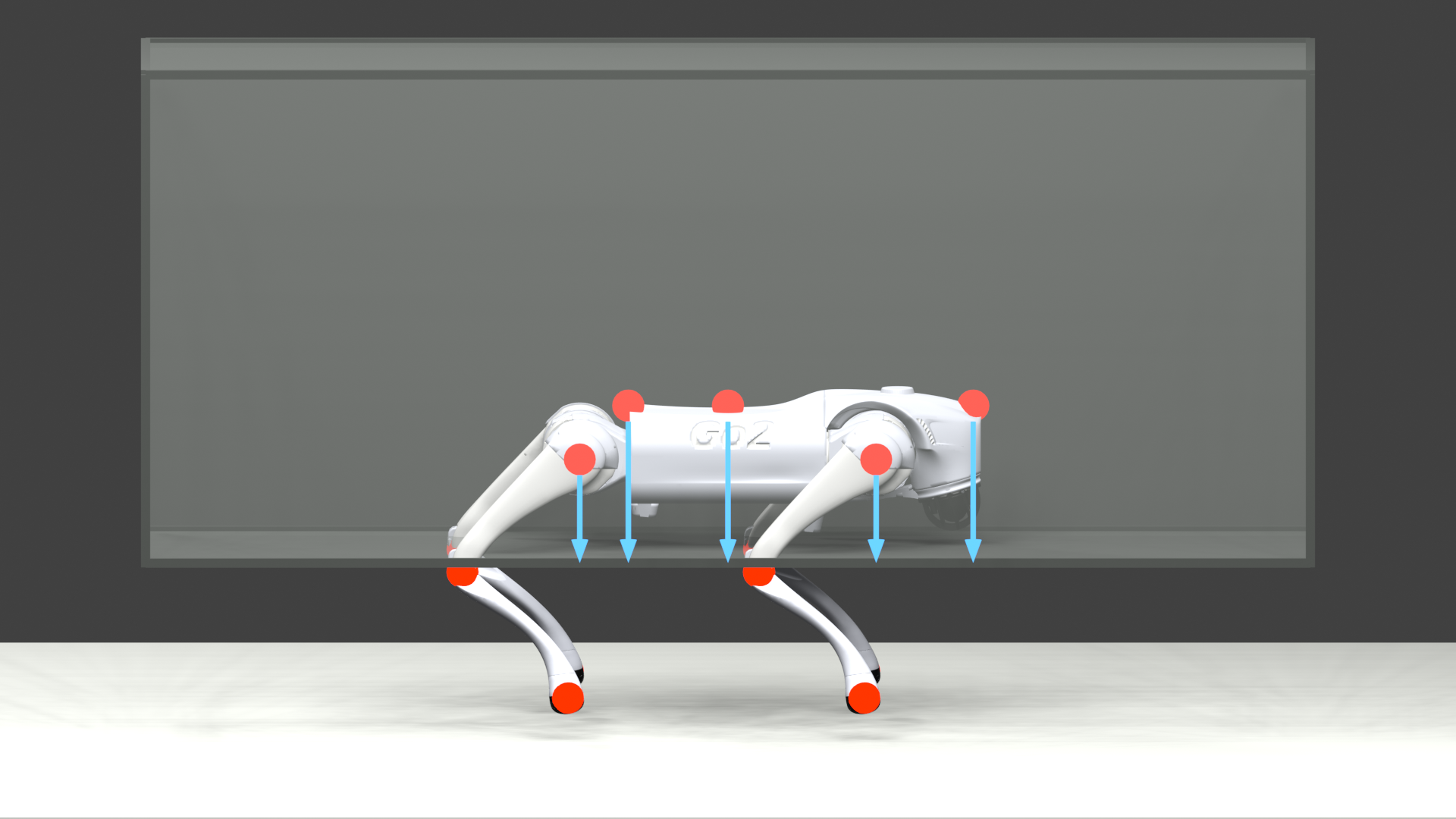}
    \hfill
    \includegraphics[width=0.45\linewidth,keepaspectratio]{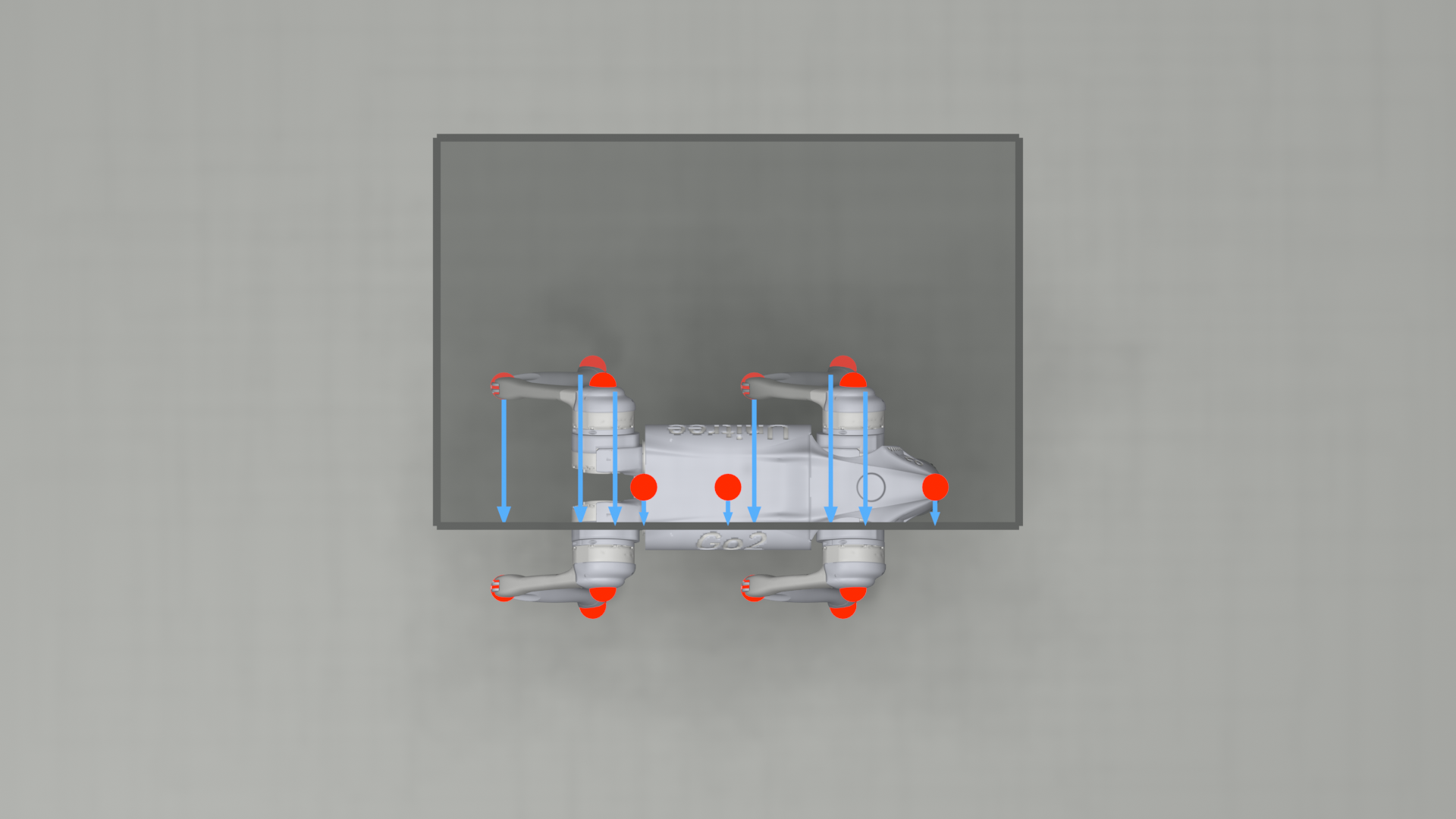}
    \caption{Visualization of the selected body keypoints and the soft penetration penalty from the side (left) and top (right) views. Red markers denote the selected keypoints, and blue lines indicate their penetration depths.}
    \label{fig:sm_keypoints}
\end{figure}

\noindent\textbf{Baseline implementation.}
The proximity baseline uses the obstacle-proximity computation of Miki et al.~\cite{miki2024learning} without terminating episodes upon obstacle collision. The SDF baseline follows Stanley et al.~\cite{stanley2026distances} and uses signed distance to the obstacle surface. Baseline and component-ablation results are reported in Fig.~\ref{fig:success} of the main paper.

\subsection{Gradient Analysis}
\label{sm:gradients}

The contact-model and obstacle-term gradient magnitudes are reported in Fig.~\ref{fig:gradient} of the main paper. Here, we examine the effect of velocity stop-gradient on position derivatives and forward motion.

\noindent\textbf{Velocity stop-gradient mechanism.}
To inspect the obstacle contribution, only the obstacle-interaction term $L_{\mathrm{obs}}$ in the SAPO actor loss is backpropagated, and its derivatives with respect to robot position $(x,y,z)$ are recorded along representative rollouts. Figures~\ref{fig:sm_crawl_gradients} and~\ref{fig:sm_bypass_gradients} compare the full formulation against the variant without the velocity stop-gradient. The latter exhibits a pronounced negative longitudinal component in the reported traces.

\begin{figure}[H]
    \centering
    \includegraphics[width=0.32\textwidth]{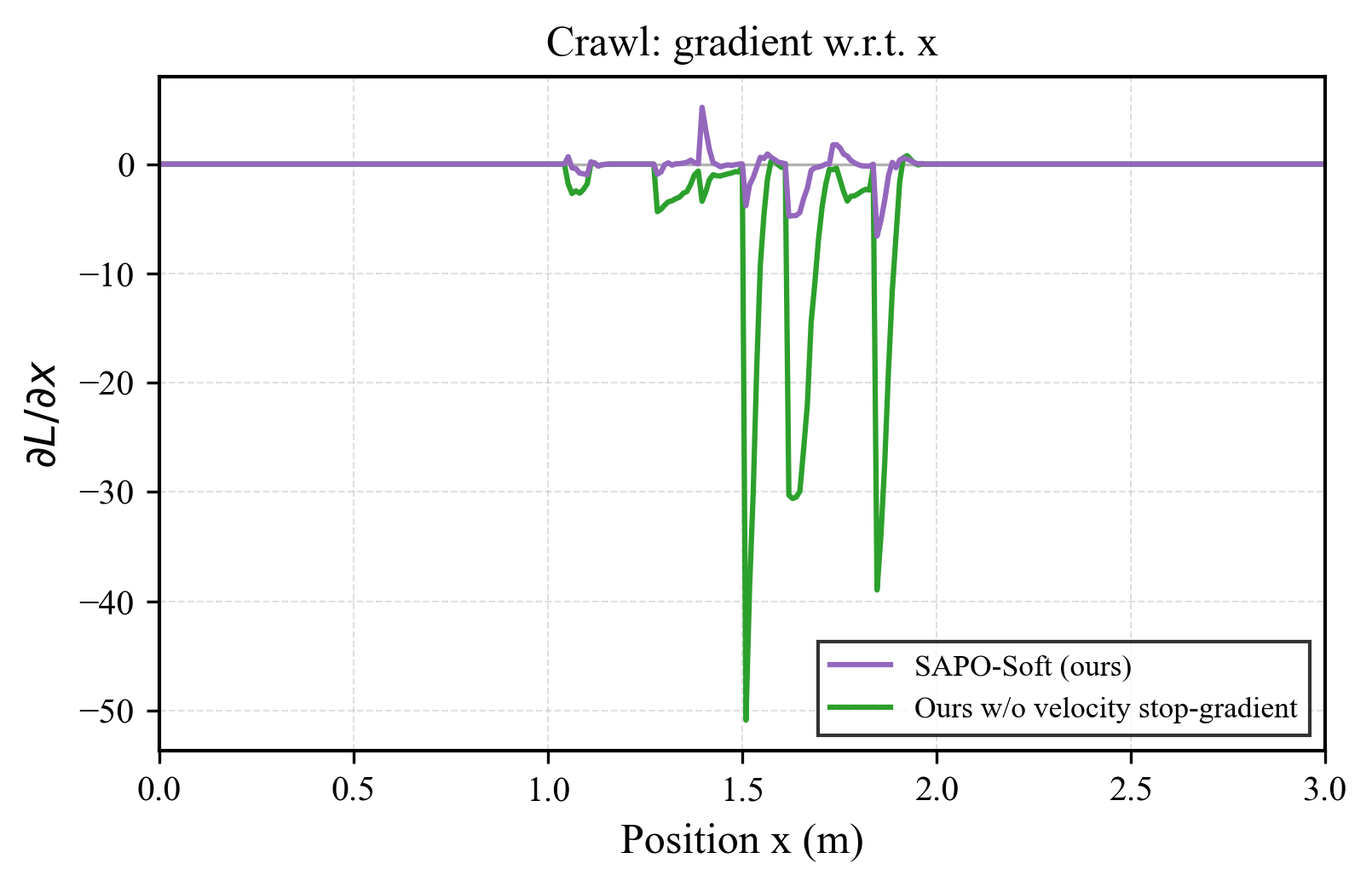}
    \hfill
    \includegraphics[width=0.32\textwidth]{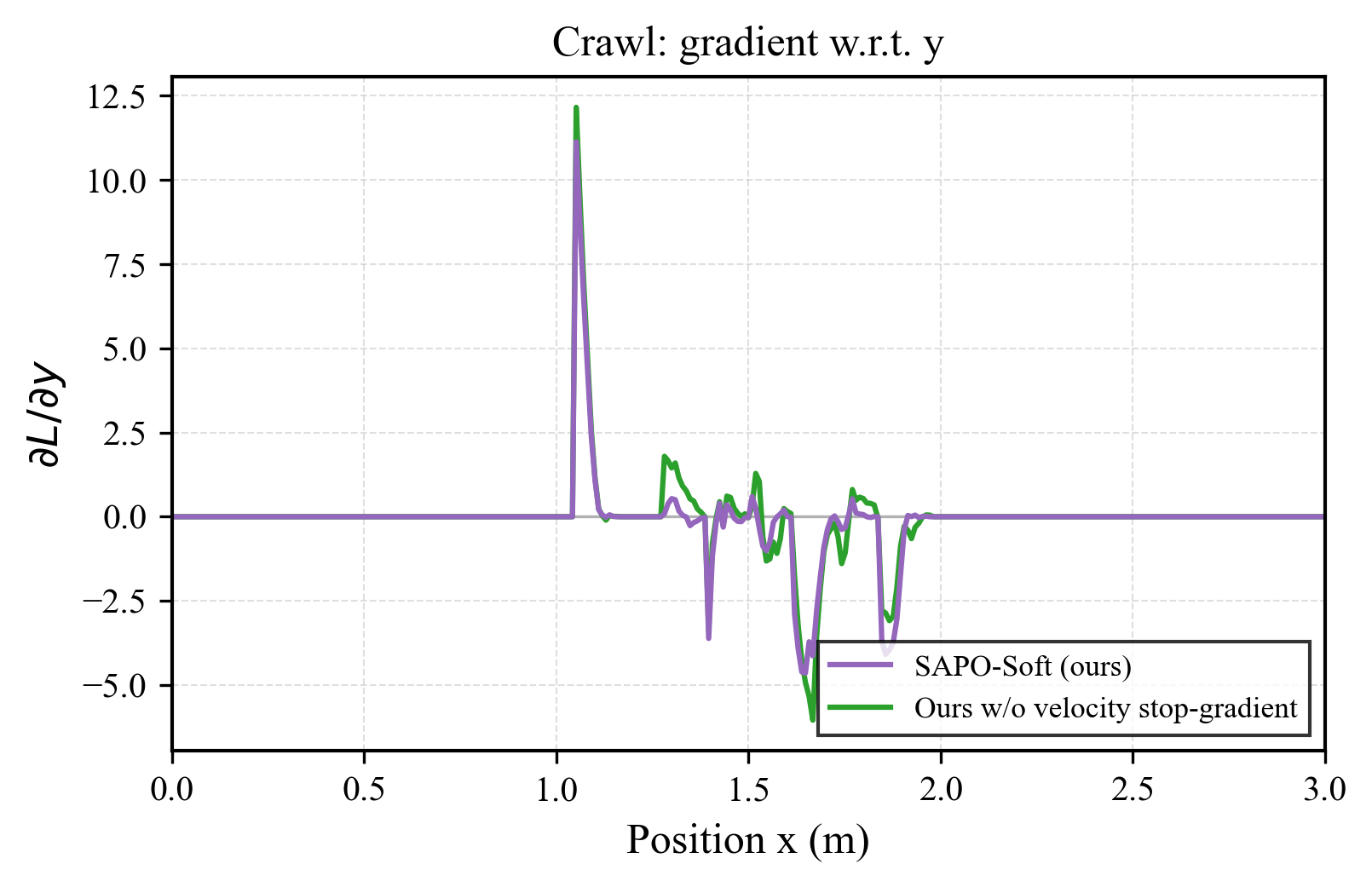}
    \hfill
    \includegraphics[width=0.32\textwidth]{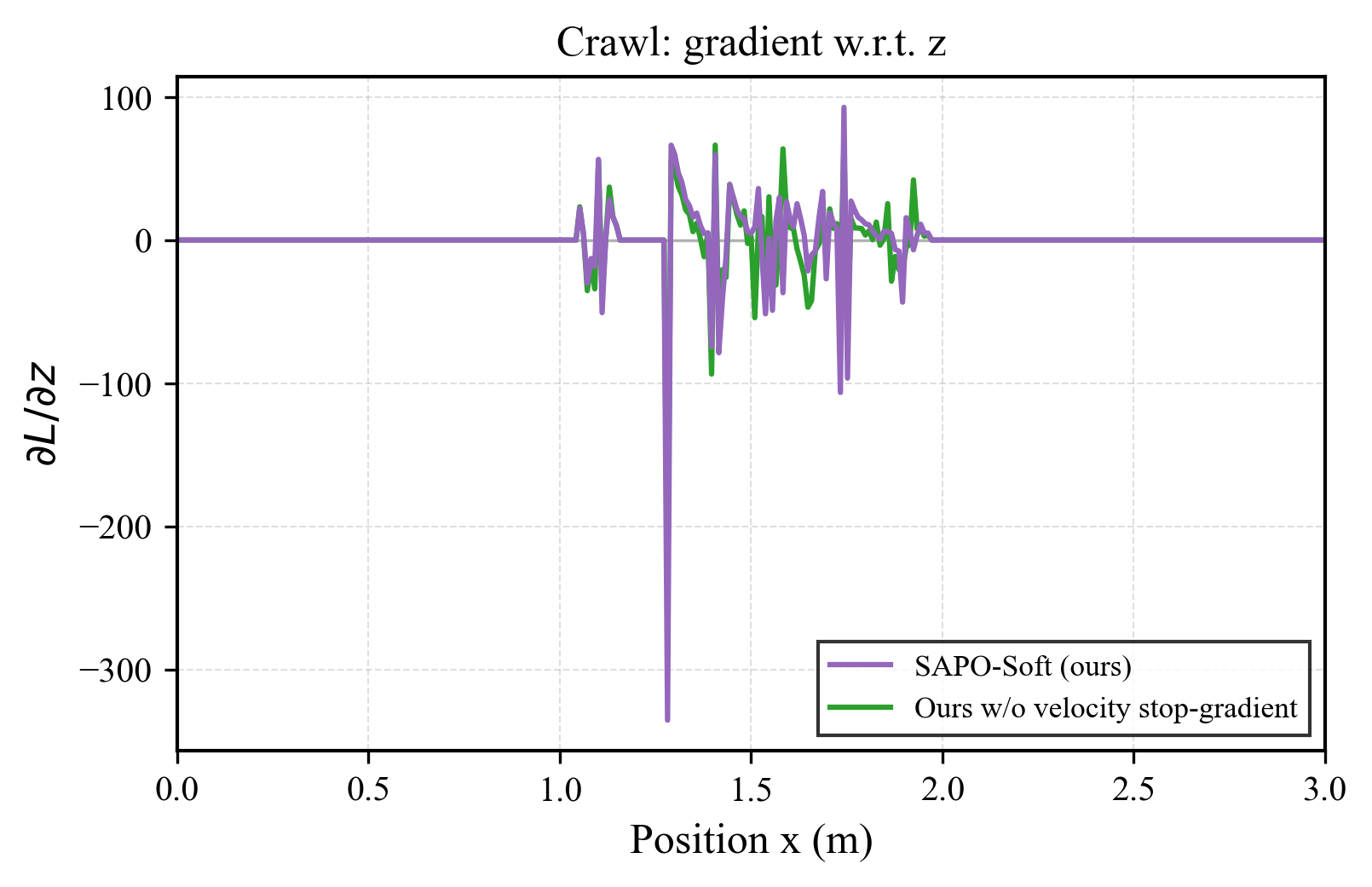}
    \caption{Obstacle-related gradients from the SAPO actor loss along a
    representative crawling rollout. From left to right, the panels show
    $\partial L_{\mathrm{obs}}/\partial x$,
    $\partial L_{\mathrm{obs}}/\partial y$, and
    $\partial L_{\mathrm{obs}}/\partial z$ as functions of longitudinal
    position. Purple denotes SAPO-Soft with velocity stop-gradient, and
    green denotes the otherwise identical formulation without it.}
    \label{fig:sm_crawl_gradients}
\end{figure}
\begin{figure}[H]
    \centering
    \includegraphics[width=0.32\textwidth]{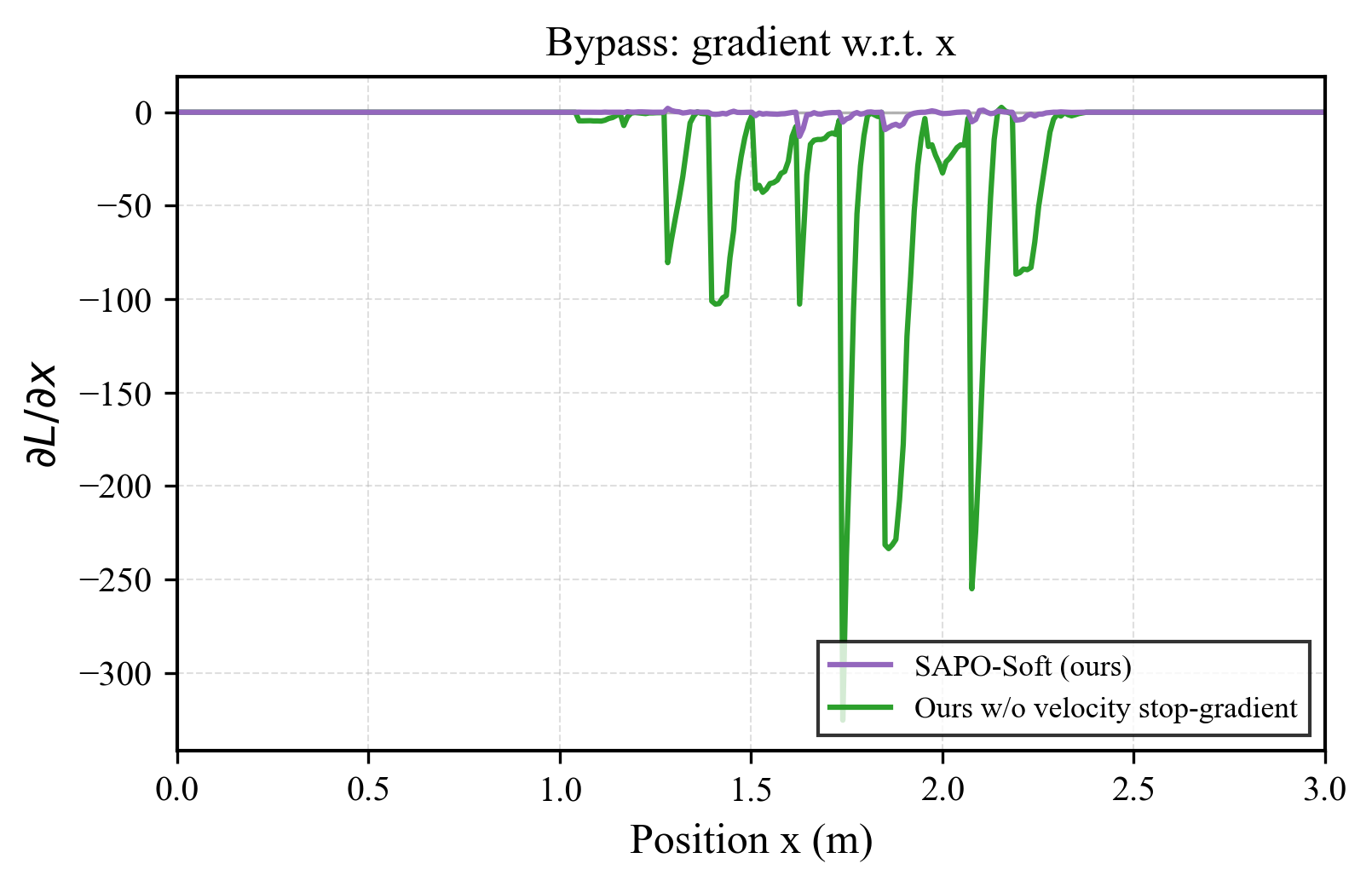}
    \hfill
    \includegraphics[width=0.32\textwidth]{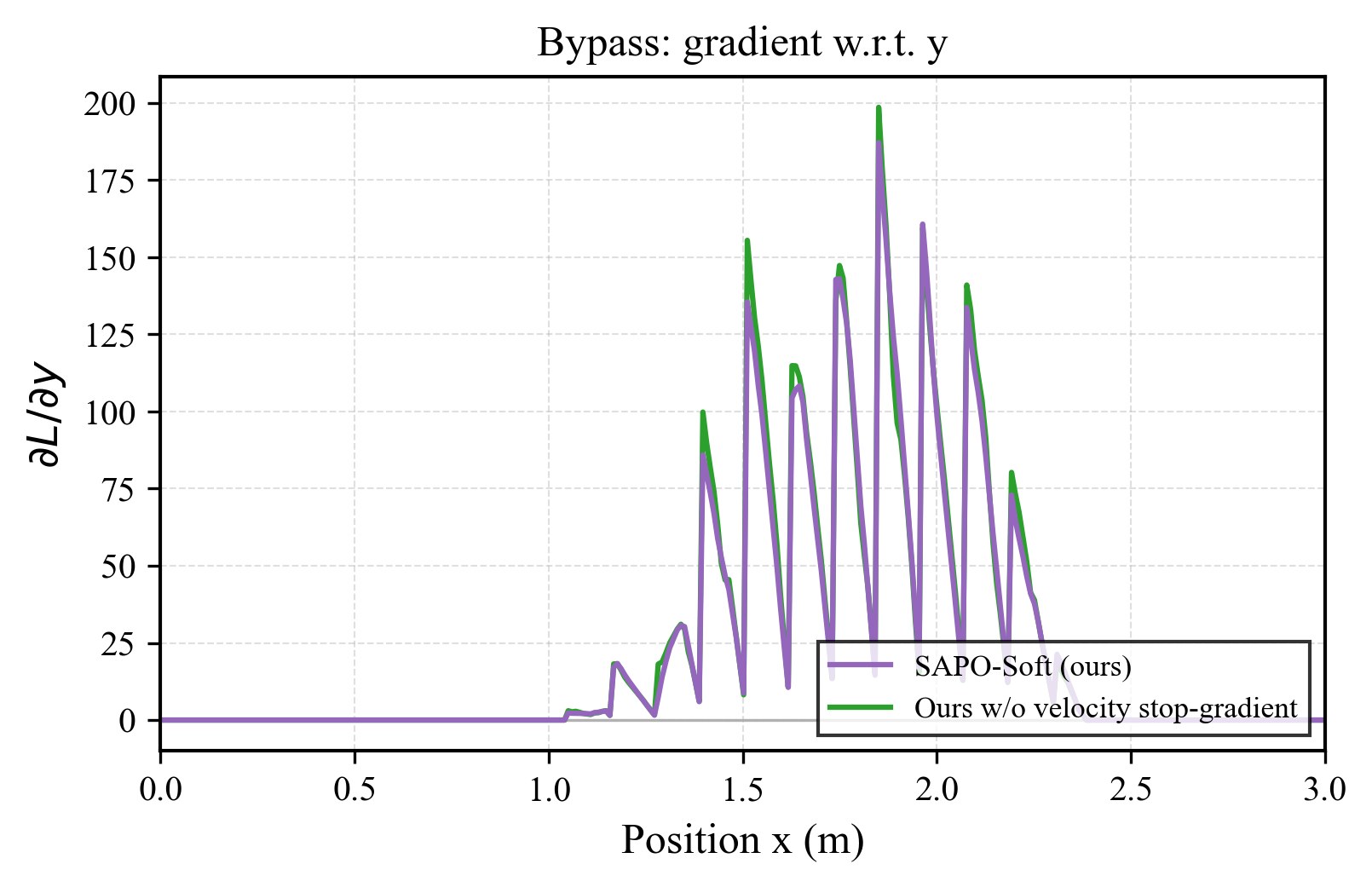}
    \hfill
    \includegraphics[width=0.32\textwidth]{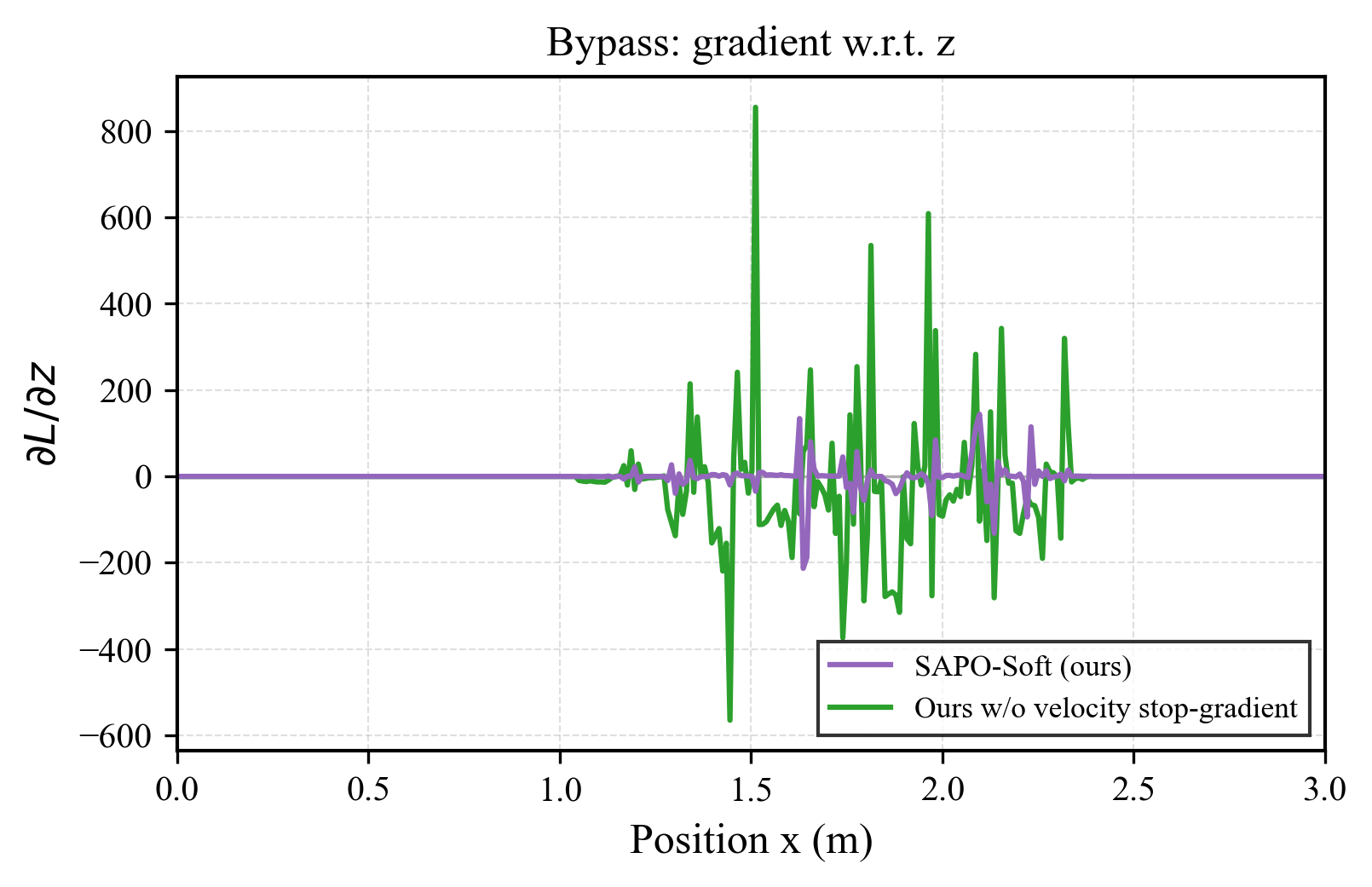}
    \caption{Obstacle-related gradients from the SAPO actor loss along a
    representative bypassing rollout. From left to right, the panels show
    $\partial L_{\mathrm{obs}}/\partial x$,
    $\partial L_{\mathrm{obs}}/\partial y$, and
    $\partial L_{\mathrm{obs}}/\partial z$. Removing the velocity
    stop-gradient introduces a dominant negative longitudinal component,
    in addition to the geometric gradients that drive lateral and vertical
    obstacle avoidance.}
    \label{fig:sm_bypass_gradients}
\end{figure}

The velocity stop-gradient in \eqref{eq:penetration_reward} removes the direct derivative through $v_x$ while retaining velocity-dependent penalty scaling and geometric feedback from penetration.

The velocity comparison in Fig.~\ref{fig:sm_stop_gradient_velocity} uses a command of $1.2\,$m/s. The overhanging obstacle has a clearance of $0.26\,$m, and the ground obstacle has a width of $0.90\,$m. The full formulation closely tracks the command and traverses both obstacles. Without the stop-gradient, velocity tracking deteriorates and traversal fails in these reported scenarios. These traces complement the gradient measurements with direct observations of forward motion.

\begin{figure}[H]
    \centering
    \includegraphics[width=0.45\linewidth,keepaspectratio]{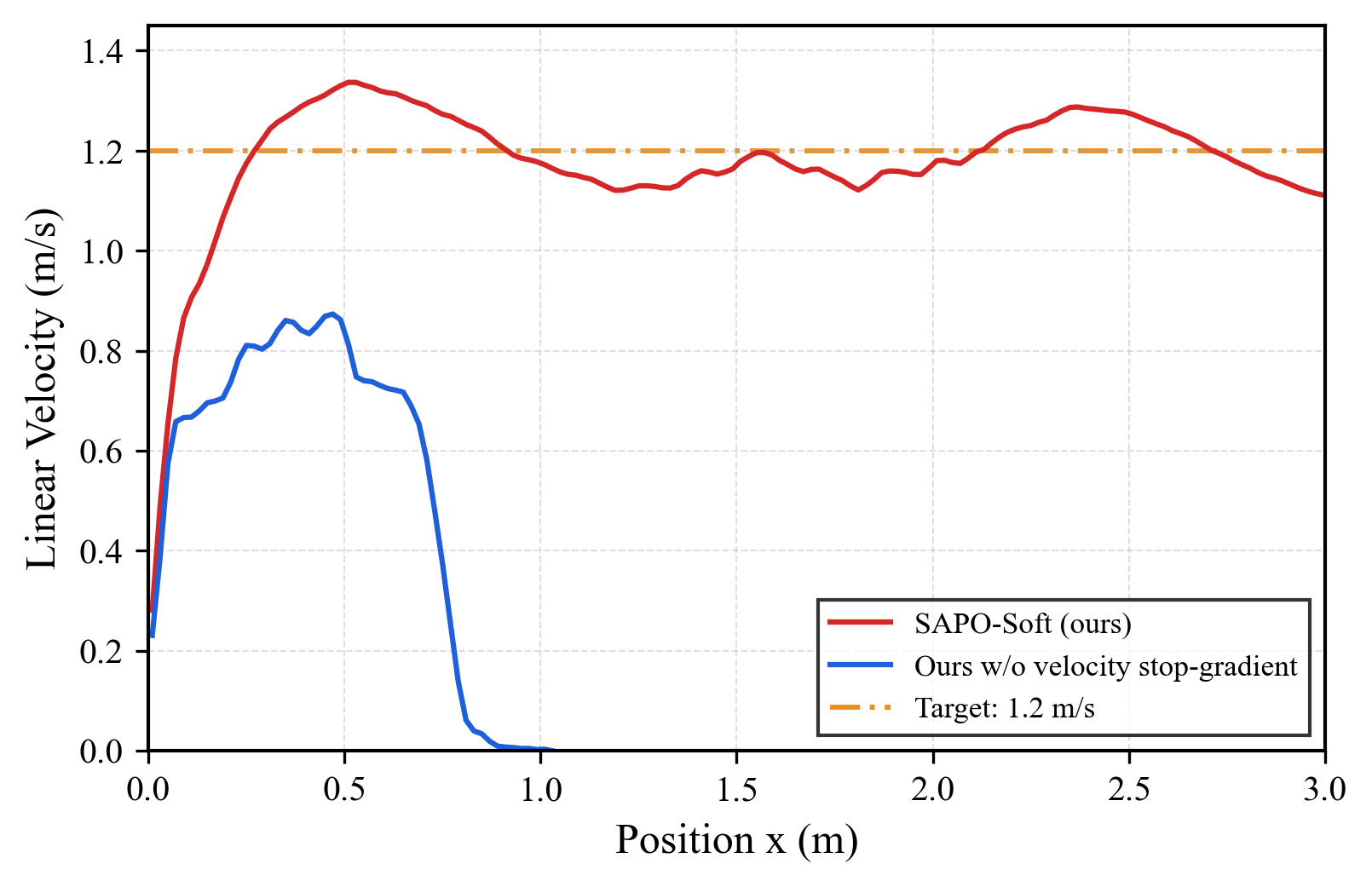}
    \hfill
    \includegraphics[width=0.45\linewidth,keepaspectratio]{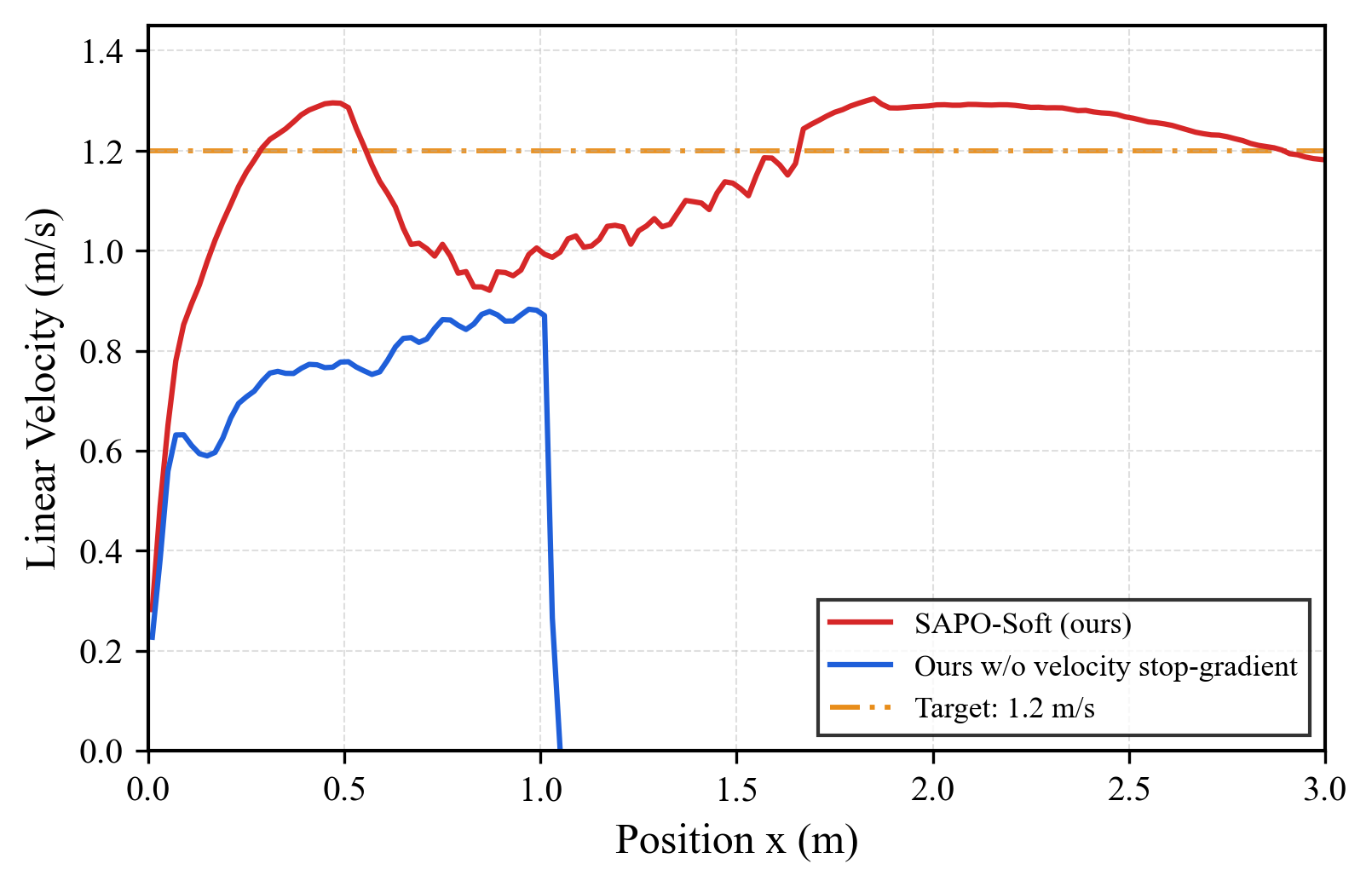}
    \caption{Forward-velocity tracking with and without the velocity
    stop-gradient for an overhanging obstacle with a clearance height of
    $0.26\,$m (left) and a ground obstacle with a width of $0.90\,$m (right).
    The commanded forward velocity is $1.2\,$m/s in both scenarios.}
    \label{fig:sm_stop_gradient_velocity}
\end{figure}

\subsection{Teacher--Student Transfer and Gating}
\label{sm:transfer}

Figure~\ref{fig:sm_distillation} compares the privileged SAPO-Soft teacher with the depth-based student. For ground obstacles, the student matches the teacher for widths up to $0.9\,$m, with a gap appearing near the feasibility boundary. For overhanging obstacles, both reliably traverse clearances from $0.33$ down to $0.28\,$m. Student performance decreases at tighter clearances from $0.27$ down to $0.24\,$m, whereas the teacher continues to pass reliably down to $0.24\,$m. Distillation therefore preserves avoidance over most of the tested range, with losses concentrated near the most restrictive geometries.

\begin{figure}[H]
    \centering
    \includegraphics[width=0.45\linewidth,keepaspectratio]{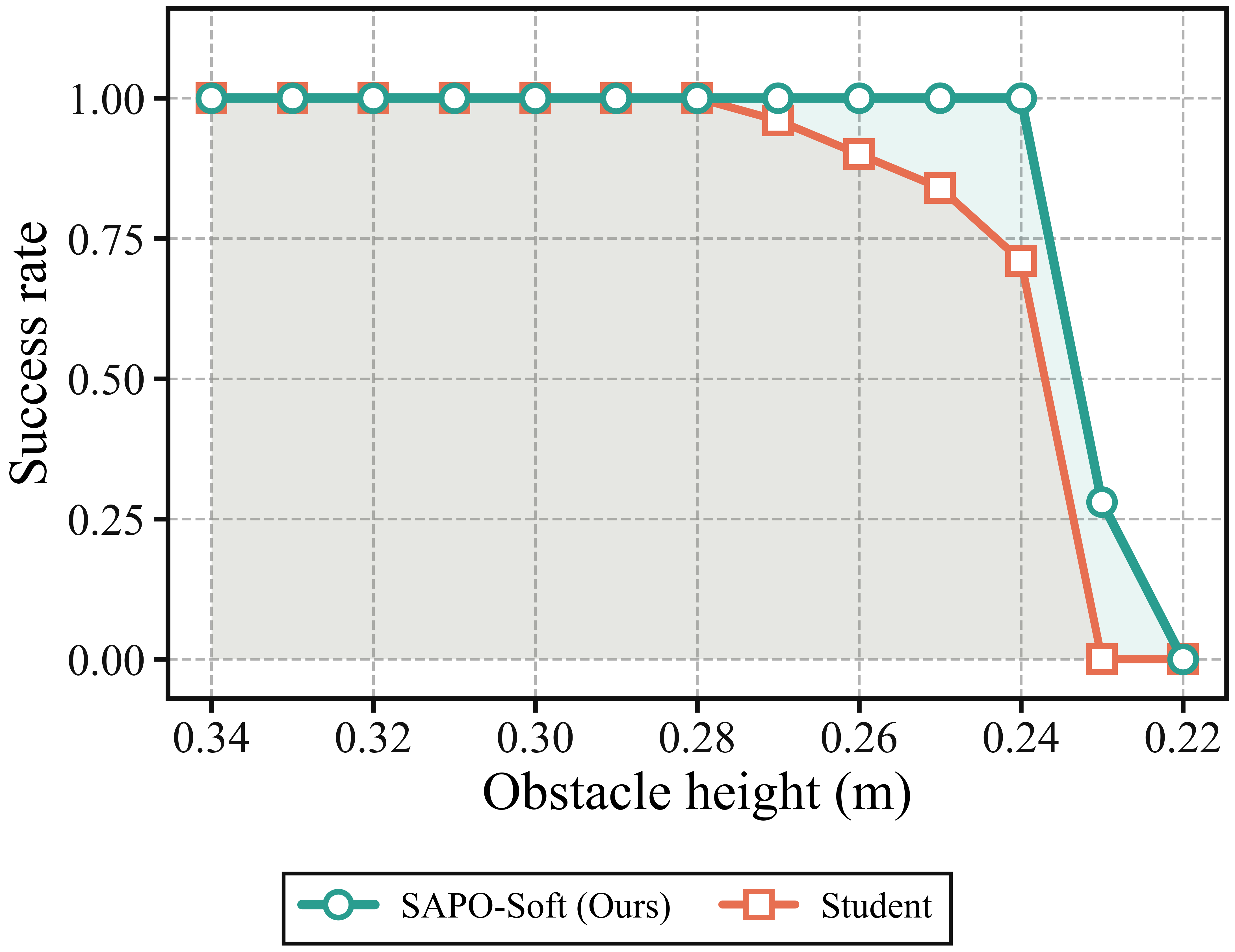}
    \hfill
    \includegraphics[width=0.45\linewidth,keepaspectratio]{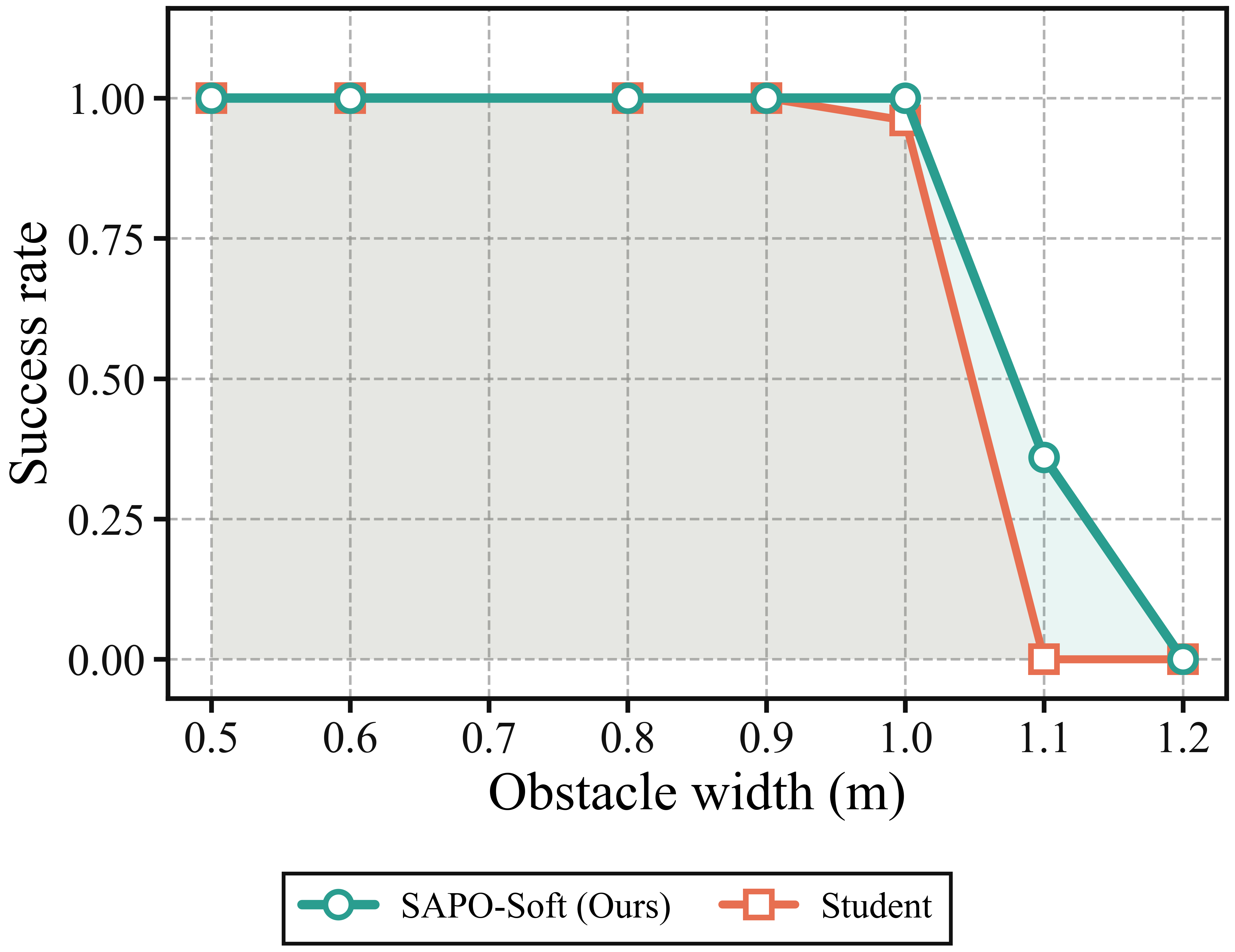}
    \caption{Comparison of obstacle-avoidance success rate between the teacher and the student for overhanging obstacles (left) and ground obstacles (right).}
    \label{fig:sm_distillation}
\end{figure}

The complementary gating comparison in Fig.~\ref{fig:vel-tracking} of the main paper shows improved velocity tracking in obstacle-free segments. The teacher--student success curves here quantify the remaining avoidance losses near the feasibility boundary.

\subsection{Optimization and Training Efficiency}
\label{sm:efficiency}

\noindent\textbf{Environment steps and wall-clock time.}
Table~\ref{tab:sm_convergence} reports three convergence criteria. The reward-plateau criterion identifies the checkpoint after which discounted reward ceases to increase consistently and fluctuates within a narrow range. Two additional criteria use reward stably above 90 and a plateau in obstacle-avoidance success rate. All listed wall-clock measurements use the same hardware and training configuration.

\begin{table}[H]
\centering
\caption{Environment steps and wall-clock time required by SAPO and PPO to reach convergence under three convergence criteria. Wall-clock times were measured on the same hardware and training configuration.}
\label{tab:sm_convergence}
\renewcommand{\arraystretch}{1.12}
\begin{tabular}{lrrrr}
\toprule
Convergence criterion & \multicolumn{2}{c}{Steps (million)} & \multicolumn{2}{c}{Time (min)} \\
\cmidrule(lr){2-3}\cmidrule(lr){4-5}
 & SAPO & PPO & SAPO & PPO \\
\midrule
Reward plateau & 17.0 & 53.0 & 366 & 130 \\
Reward stably above 90 & 10.5 & 60.0 & 226 & 147 \\
Avoidance success-rate plateau & 16.0 & 67.0 & 344 & 166 \\
\bottomrule
\end{tabular}
\end{table}

The avoidance-success plateau is the selected criterion because it directly measures the task objective. Under this criterion, SAPO requires 16M environment steps and PPO requires 67M, corresponding to approximately $4.2\times$ fewer samples for SAPO. Their training times are 344 and 166 minutes, respectively, so SAPO takes approximately $2.1\times$ as long in wall-clock time. The sample-efficiency improvement therefore does not imply faster training in elapsed time.

\subsection{Parameter Sensitivity and Simulation Settings}
\label{sm:parameters}

\noindent\textbf{Penalty parameters.}
The velocity-shaping parameter uses $\sigma=0.50$, and the conditional negative constant is $k=-0.3$. To distinguish that constant from the task-specific penetration weights, the latter are denoted by uppercase $K_{\mathrm{overhanging}}$ and $K_{\mathrm{ground}}$ throughout this supplement. Their base values are 40 and 10, respectively.

Table~\ref{tab:sm_penalty_sensitivity} reports one-parameter-at-a-time sweeps with the remaining configuration fixed, using the 90\% success threshold defined in Section~\ref{sm:protocol}. The results show that increasing the penalty weight does not uniformly improve the obstacle limit.

\begin{table}[H]
\centering
\caption{Sensitivity to $\sigma$ and the task-specific penetration weights $K$. Each obstacle limit is the most challenging geometry at which success is at least 90\%. Bold entries indicate base settings; n/a denotes a metric not applicable to the task-specific sweep.}
\label{tab:sm_penalty_sensitivity}
\renewcommand{\arraystretch}{1.12}
\begin{tabular}{llcc}
\toprule
Parameter & Tested value & \shortstack{Min. overhanging-obstacle\\height (m)} &
\shortstack{Max. ground-obstacle\\width (m)} \\
\midrule
\multirow{3}{*}{$\sigma$}
 & 0.25 & 0.29 & 0.90 \\
 & \textbf{0.50 (base)} & \textbf{0.24} & \textbf{1.00} \\
 & 0.75 & 0.25 & 1.00 \\
\midrule
\multirow{3}{*}{$K_{\mathrm{overhanging}}$}
 & 20 & 0.25 & n/a \\
 & \textbf{40 (base)} & \textbf{0.24} & n/a \\
 & 60 & 0.24 & n/a \\
\midrule
\multirow{3}{*}{$K_{\mathrm{ground}}$}
 & 5 & n/a & 0.70 \\
 & \textbf{10 (base)} & n/a & \textbf{1.00} \\
 & 15 & n/a & 0.90 \\
\bottomrule
\end{tabular}
\end{table}

\noindent\textbf{Contact and friction.}
Table~\ref{tab:sm_contact_parameters} gives the robot and ground settings used in simulation, together with their effective pair averages. These contact dynamics are shared by the PPO, SHAC, and SAPO comparisons.

\begin{table}[H]
\centering
\caption{Contact and friction parameters used in all simulation experiments.
}
\label{tab:sm_contact_parameters}
\renewcommand{\arraystretch}{1.12}
\begin{tabular}{llccc}
\toprule
Parameter & Unit & Robot geometry & Ground & \shortstack{Effective\\(pair average)} \\
\midrule
$k_n$ (\texttt{contact\_ke}) & N/m & $4.0\times10^{3}$ & $1.0\times10^{5}$ & $5.2\times10^{4}$ \\
$k_d$ (\texttt{contact\_kd}) & N$\cdot$s/m$^2$ & $1.0\times10^{3}$ & $1.0\times10^{3}$ & $1.0\times10^{3}$ \\
$k_t$ (\texttt{contact\_kf}) & N$\cdot$s/m & $3.0\times10^{2}$ & $1.0\times10^{3}$ & $6.5\times10^{2}$ \\
$\mu$ (\texttt{contact\_mu}) & dimensionless & $1.0$ & $1.0$ & $1.0$ \\
\bottomrule
\end{tabular}
\end{table}

\noindent\textbf{Entropy regularization.}
The entropy coefficient $\alpha$ is tuned automatically against a target entropy; it is not fixed throughout training. Table~\ref{tab:sm_entropy} lists its initialization, optimizer, target, and observed values. The curriculum and episode termination settings are given in Section~\ref{sm:protocol}.

\begin{table}[H]
\centering
\caption{Configuration and observed values of the entropy coefficient $\alpha$.
}
\label{tab:sm_entropy}
\renewcommand{\arraystretch}{1.12}
\begin{tabular}{ll}
\toprule
Item & Value \\
\midrule
Tuning scheme & Automatic (target-entropy based) \\
Initial value $\alpha_0$ & $0.01$ \\
Target entropy & $-0.6$ \\
Optimizer for $\alpha$ & AdamW, learning rate $2\times10^{-3}$ \\
Observed range during training & $0.006$ to $0.02$ \\
Observed median during training & $\approx 0.01$ \\
\bottomrule
\end{tabular}
\end{table}

\clearpage
\endgroup

\end{document}